%% file: main_00.tex
\documentclass[journal]{IEEEtran}
\usepackage{amssymb}
\usepackage{graphicx}
\usepackage{amsmath}
\usepackage{url}
\usepackage{booktabs}
\usepackage{threeparttable}
\usepackage{subcaption}
\usepackage{xcolor}
\usepackage{algorithm}
\usepackage{algpseudocode}   % algorithmicx
\usepackage{afterpage}
\usepackage[table]{xcolor}
\newcommand{\orikey}[2]{%
  \raisebox{0.35ex}[0pt][0pt]{%
    \rotatebox[origin=c]{#1}{\textcolor[HTML]{#2}{\rule{8pt}{1.3pt}}}}%
}

\ifCLASSINFOpdf
\else
\fi
\begin{document}

%%%%%% TITLES
%\title{Fingermark Feature Extraction and Quality Assessment: A Recognition-Driven Unified Multi-Task Framework}
%\title{Efficient Fingermark Representation Learning: \\ A Unified Multi-Task Framework}
%\title{Efficient Fingerprint Representation Learning, Feature Extraction, and Quality: \\ A Unified Multi-Task Framework}
%\title{AFID: A Multi-Task Forensic Framework for Automated Fingerprint Identification, Quality Assessment and Feature Extraction}
\title{AFID: A Unified Open Framework for Automated Fingermark Identification, Quality Assessment and Feature Extraction}
%\title{Efficient Fixed-Length Fingermark Representation with Feature Extraction and Quality Assessment Multi-task decoders}


\author{Tim Oblak,
        Rudolf Haraksim,
        Peter Peer% <-this % stops a space
        
\thanks{T. Oblak and P. Peer are with the University of Ljubljana, Faculty of Computer and Information Science in Ljubljana, Slovenia, e-mail: \{tim.oblak,peter.peer\}@fri.uni-lj.si}% <-this % stops a space
\thanks{R. Haraksim is with the Joint Research Centre of the European Commission in Ispra, Italy, e-mail: rudolf.haraksim@ec.europa.eu}% <-this % stops a space
%\thanks{Manuscript received April 19, 2005; revised September 17, 2014.}
}


\markboth{Journal of \LaTeX\ Class Files,~Vol.~13, No.~9, September~2014}%
{Shell \MakeLowercase{\textit{et al.}}: Bare Demo of IEEEtran.cls for Journals}



\maketitle

\begin{abstract}
Automated fingermark identification is the foundation of forensic investigation, yet progress in the field is held back by fragmented, closed-source solutions trained on private or discontinued data. We present AFID, a unified open-source framework for friction ridge image processing that performs recognition, quality assessment, and feature extraction based on a single shared encoder, trained exclusively on publicly available data. At its core is a fixed-length representation learned for identity discrimination, trained under heavy augmentation. Despite applying essentially no preprocessing beyond resizing and padding at inference, AFID sets a new state of the art in fixed-length fingermark recognition, leading identification across NIST~SD~27 ($\textbf{67.6\%}$ rank-1) , SD~302 ($\textbf{54.9\%}$ rank-1), and SD~303 ($\textbf{67.6\%}$ rank-1), surpassing a commercial matcher on fingermarks. From the same frozen backbone, a quality assessment module predicts recognition utility more accurately than any compared baseline and generalizes across independent matchers, while lightweight decoders recover minutiae, ridge orientation, and segmentation competitive with dedicated methods. The framework proves that a single, efficiently trained encoder can support the full fingermark processing pipeline, from recognition through quality assessment all the way to feature extraction. To accelerate research on fingermark analysis even further, we release the code, models, and annotations to the community.
\end{abstract}

\begin{IEEEkeywords}
fingermark, latent fingerprint, identification, quality assessment, feature extraction, forensics
\end{IEEEkeywords}

% \fi
\IEEEpeerreviewmaketitle

\input{main_01_introduction}

% This was in the template after the introduction, not sure why. 
%\hfill September 17, 2014

\input{main_02_related_work}

\input{main_03_methods}

\input{main_04_experiments}

\input{main_05_conclusion}
\ifCLASSOPTIONcaptionsoff
  \newpage
\fi



\bibliographystyle{IEEEtran}
\bibliography{bib.bib}


%\end{IEEEbiography}


%\vfill


\input{main_06_supplementary}

\end{document}

%% file: main_01_introduction.tex
\begin{figure*}[t]
  \centering
  \includegraphics[width=\textwidth]{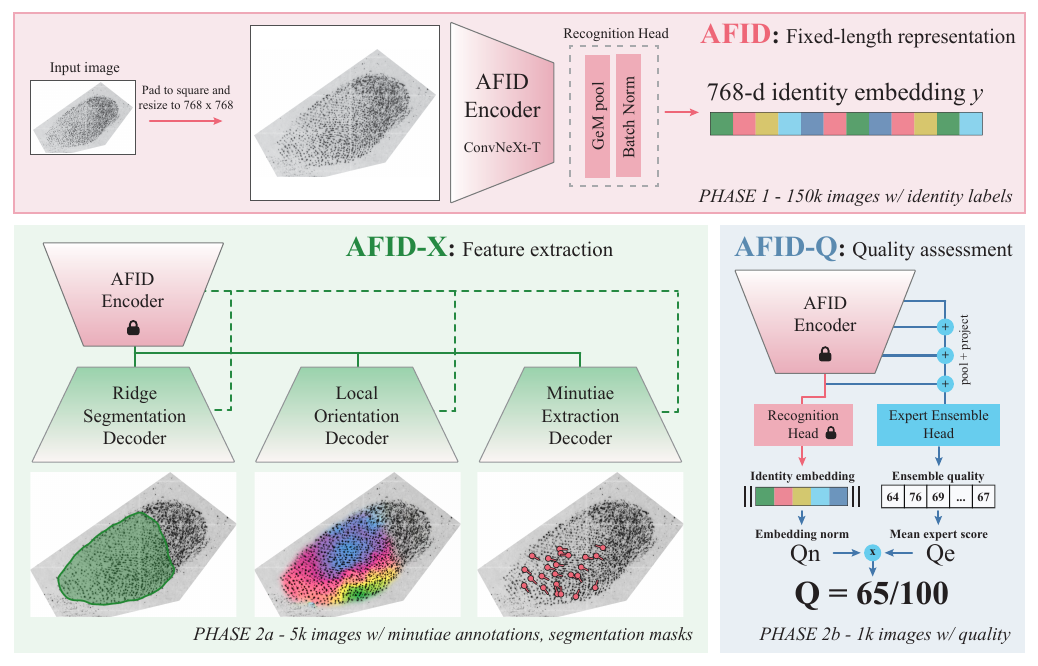}
  \caption{Overview of the AFID framework. \textbf{Phase 1} trains the AFID encoder (ConvNeXt-T) for recognition, producing a fixed-length 768-dimensional identity embedding from a minimally preprocessed input (pad to square, resize to $768 \times 768$). \textbf{Phase 2} freezes the encoder and attaches lightweight task heads on top of it: AFID-X decoders recover ridge segmentation, local orientation, and minutiae (Phase~2a), while AFID-Q derives a quality score by fusing the embedding norm $Q_n$ with the mean output of an expert-supervised regression head $Q_e$ (Phase~2b). The decreasing annotation requirement across phases ($150$k identity labels, $5$k feature annotations, $1$k quality labels) reflects the data efficiency of the shared-encoder design.}
  \label{fig:overview}
\end{figure*}

\section{Introduction}
%1. Fingermarks in forensics. ACE-V framework. identification pipeline is a joint effore between both humans and algorithms.

\IEEEPARstart{F}{riction} ridge impressions are among the most widely used forms of trace evidence in forensic investigation. When such impressions are recovered from a crime scene, typically as residues left on a surface and developed through chemical or physical processing, they are referred to as \emph{fingermarks}.\footnote{The terms \emph{fingermark} and \emph{latent fingerprint} are used largely interchangeably in the literature. We adopt \emph{fingermark} throughout, following European forensic practice and recent ISO/IEC standardization efforts.} Because friction ridge patterns are effectively unique and persistent, a fingermark recovered from a scene can be compared against reference impressions to establish or exclude a source, making it a foundational backbone evidence in criminal investigations~\cite{barnes2010fingerprint}. At scale, this comparison is performed by Automated Fingerprint Identification Systems (AFIS) that search a probe fingermark against galleries of millions of references and return a ranked candidate list for a trained fingerprint expert to examine. Fingermarks, being partial, distorted, and often noisy impressions, 
are difficult to process and the burden of resolving ambiguous or degraded samples falls heavily on the examiner. The throughput and accuracy of the automated search therefore directly shape examiner workload, and improving it is a step toward fully automated, high-volume forensic identification.

Historically, friction ridge recognition has relied on features defined by domain expertise, most prominently minutiae points. Automated systems built on such features have matured over decades and remain the operational standard. Yet handcrafted features restrict a recognition system to a predetermined subset of the information present in an impression. Modern computer vision has largely moved in the opposite direction, replacing engineered features with representations learned directly from data. Learned representations consistently capture identity more effectively than handcrafted descriptors, as evident in the field of face recognition~\cite{deng2019arcface,meng2021magface}. Beyond their discriminative power, learned representations offer a decisive practical advantage. The fixed-length embeddings such models produce simplify comparison to a distance computation between two vectors, which is orders of magnitude faster than matching variable-size minutiae sets. At the scale of national and international AFIS deployments, the difference between a vector comparison and a graph-matching operation translates into a large difference in search time. 

These advantages, however, have been slower to reach the fingerprint domain and the reason can be largely attributed to lack of data. Face recognition has been driven by large datasets containing millions of identities, whereas fingerprint data is comparatively scarce and difficult to collect. The situation was worsened by the withdrawal of several previously public datasets. A strong representation must therefore be learned from far less data than face recognition has had at its disposal.

The scarcity of data is mirrored by a scarcity of open-source solutions. In contrast to face recognition, where strong pretrained models are widely available, fingerprint recognition offers comparatively few. Many published methods are never released, and those that are, often depend on training data that is no longer accessible: several influential systems were developed on datasets that are now withdrawn, such as NIST SD4~\cite{watson1992sd4}, SD14~\cite{watson2001sd14}, and SD27~\cite{sd27}, or on private operational datasets that were never public in the first place. A method trained on such data cannot be retrained or independently verified, and one released without weights cannot be run at all. The result is a field, in which direct comparison between methods is difficult and cumulative progress is slow, as each new approach is evaluated under its own protocol against baselines that cannot always be reproduced.

These difficulties became evident in the ongoing effort to standardize fingermark quality assessment. Sample quality, a prediction of how useful an impression will be for recognition, governs core operational decisions such as whether a mark is worth submitting to an AFIS and how examiner attention is prioritized during a forensic investigation. For plain and rolled fingerprints, this need is met by NFIQ~2~\cite{tabassi2021nfiq2}, the open, standardized reference implementation of ISO/IEC~29794-4~\cite{iso29794-4:2024}, widely deployed across civil and government systems. No comparable resource exists for fingermarks, which occupy a far more difficult regime: partial, distorted, and recovered from uncontrolled surfaces. The ISO/IEC~29794-12 standard is being developed to address this, but it lacks the open reference implementations, benchmarks, and evaluation protocols, particularly for quality assessment and feature extraction, that would let the community build on a common foundation.

In this work we address these gaps with AFID, a unified and fully open framework for friction ridge image processing, shown in Figure~\ref{fig:overview}. Rather than extracting features and then matching as conventional pipelines do, AFID first learns a recognition representation and then derives feature extraction and quality assessment from it. A single encoder, trained for recognition on publicly available data only, serves as the shared foundation for all downstream tasks. Our contributions are as follows:

\begin{itemize}
    \item \textbf{A data-efficient fixed-length representation for friction ridge images.} We train a single encoder for fingerprint and fingermark recognition, using only publicly available data. The encoder produces fixed-length embeddings that are robust to rotation and scale without any explicit alignment. The model requires essentially no preprocessing beyond resizing and padding. The resulting model sets a new state-of-the-art in fixed-length fingermark recognition, leading identification across major fingermark datasets, surpassing even a commercial matcher.

    \item \textbf{A recognition-aligned quality metric.} From the same frozen representation, we derive a quality metric, that predicts recognition utility more accurately than existing methods and generalizes across independent matchers rather than acting as a matcher-specific confidence score.

    \item \textbf{Feature extraction from the shared representation.} We attach lightweight decoders to the frozen encoder to recover the core fingermark feature set, minutiae, ridge orientation, and ridge segmentation, at a level competitive with dedicated methods. We demonstrate that the learned representation retains the local ridge structure from which these features derive.

    \item \textbf{A full open release.} We publicly release the framework, the pretrained models, and the annotations produced during this work, providing a reproducible foundation for a field that has lacked one and a candidate reference implementation for the emerging ISO/IEC~29794-12 standard.
\end{itemize}

%% file: main_02_related_work.tex
% SECTION STATUS: rougly includes most of the imporstant related work, taxonomy is there, but needs some more work
% - probably needs to be trimmed down a bit, 
% - for each paragraph, make sure that we have a good summary and how our appraoch relates to it 
% - also need to finish the final "feature extration 
% - some references might be more suitable in other categories

\section{Related Work}
\label{sec:related}
Recognition, feature extraction, and quality assessment for friction ridge images are closely connected: all three operate on mostly overlapping feature sets derived from ridge structure. %This connection is commonly used in the literature, with minutiae, orientation, and segmentation frequently being incorporated into recognition and quality pipelines as auxiliary signals. 
%The following section surveys the prior work along these three topics.

\subsection{Representation Learning}
\label{sec:rw:representation}
Modern fingerprint recognition methods fall into three broad families. \emph{Minutiae-based methods} follow the classical AFIS approach. They often apply several enhancing and preprocessing steps and ultimately rely on minutiae-based matching. These systems achieve the highest accuracy on difficult fingermark datasets but are computationally heavy and multi-stage. MSU-LatentAFIS~\cite{cao2019automated,cao2020endtoend} is a representative open example, and the recent LFR-Net~\cite{grosz2023lfrnet} reaches $84.11\%$ rank-1 on SD~27 by fusing local minutiae descriptors with a global embedding. \emph{Comparison-based methods}~\cite{he2022pfvnet,guan2024joint,qiu2024ifvit} instead learn to match image pairs directly, producing accurate correspondences but requiring both the probe and the gallery images to be re-processed for each comparison, which does not scale to large-gallery search.

The third family, producing \emph{fixed-length global embeddings}, is the focus of this work. Each impression is mapped to a single compact representation. This reduces matching to a distance computation and enables fast search at scale. DeepPrint~\cite{engelsma2021deepprint} established the approach by forming the fixed representation based on minutiae and texture information. It also introduced the localization module which first aligns the impression. The biggest advance in this domain was the adoption of angular-margin losses from face recognition. AFR-Net~\cite{grosz2024afrnet} combined a convnet and a Vision Transformer (ViT) with ArcFace~\cite{deng2019arcface} loss, the largest single contributor to its performance. More recently, Pan et al.~\cite{pan2024fdd} proposed fixed-length dense descriptors, trained using CosFace~\cite{wang2018cosface}. The authors extended the approach in FLARE~\cite{pan2026flare} with pose-based alignment and additional input enhancement. Limitations are common across these methods. Most were never released, though FLARE is a notable exception, and most are trained on large corpora of proprietary or discontinued data. Many also depend on an explicit alignment step that estimates and corrects pose before the representation is extracted. AFID addresses these limitations directly: it trains on public data alone, using heavy augmentation and synthetic data for more data efficient training and requires no alignment at inference.

\subsection{Feature Extraction}
\label{sec:rw:features}
Minutiae points are at the core of friction ridge feature extraction. Early CNN-based extractors already reported results on fingermarks~\cite{sankaran2014latent}, with FingerNet~\cite{tang2017fingernet} jointly predicting orientation, segmentation, and minutiae in a single network. MinutiaeNet~\cite{nguyen2018minutiaenet} introduced a two-stage design, a CoarseNet producing minutiae and orientation maps followed by a FineNet that refines candidate patches. %Cao et al.~\cite{cao2019automated,cao2020endtoend} combined autoencoder enhancement, a minutiae extractor, and texture descriptors. 
More recent work favors U-Net architectures: Finger-UNet~\cite{gavas2023fingerunet} proposes enhancement, minutiae detection, and orientation in a multi-branch decoder, and Liu et al.~\cite{liu2022multitask_minutiae} learn minutiae location and direction jointly on full contactless images. Auxiliary tasks that support recognition have been addressed with similar architectures. For fingermark segmentation, Nguyen et al.~\cite{nguyen2018segfinnet} proposed SegFinNet, a fully convolutional approach that outperformed human markup on NIST~SD27. Liu and Qian~\cite{liu2021nested_unet} used nested U-Nets for joint segmentation and enhancement. For orientation estimation, approaches have progressed from dictionary-based methods~\cite{yang2014localized} to convolutional networks~\cite{cao2015latent}, with recent methods such as RefNet~\cite{duan2021refnet} using statistical priors from high-quality prints to recover orientation fields together with reliability scores.

\subsection{Quality Assessment}
\label{sec:rw:quality}
The standardized approach for fingerprint quality is NFIQ\,2~\cite{tabassi2021nfiq2}, an open-source model that conforms to ISO/IEC~29794-4~\cite{iso29794-4:2024}. Its feature set is designed for controlled acquisition and its predictive power degrades on forensic fingermarks~\cite{oblak2021afqa,oblak2022framework}. Early fingermark-specific metrics relied on handcrafted descriptors and heuristics~\cite{yoon2013lfiq,Sankaran2013,swofford2021method}, before the field shifted toward learned models~\cite{ezeobiejesi2018latent,oblak2022framework,oblak2023pafqa}.

Because expert judgment is central to forensic casework, a parallel thread has built quality metrics from examiner annotations. The FBI and Noblis developed LQMetric~\cite{kalka2020lqmetric}, trained on examiner markup and calibrated against their operational AFIS. It has been in use since 2014, but only recently released as open source (OpenLQM) by NIST. The JRC campaign~\cite{haraksim2023annotation} collected quality labels for 1{,}000 NIST SD~301 and SD~302 images, each independently assessed by ten certified examiners. Oblak et al.~\cite{oblak2023pafqa} used this data to develop pAFQA, predicting quality as a probability distribution to produce both a quality value and an uncertainty estimate. More recent work continues to rely on such expert labels~\cite{chugh2018crowd,huang2026dbfqa}, which are highly predictive but costly to obtain.

A different line of work, largely originating  from face recognition, derives quality directly from the recognition model rather than annotated labels. Magnitude-aware margin losses cause embedding norms to correlate with sample quality~\cite{meng2021magface,kim2022adaface}, and related methods estimate quality from embedding stability or class-distance geometry~\cite{terhorst2020serfiq,ou2021sddfiqa,boutros2023crfiqa}. Such recognition-derived quality is largely unexplored for friction ridge images, mostly because of the lack of open recognition models, and the public data to train them. We build on this idea, combining a norm-based signal from the recognition encoder with a supervised expert signal to produce a predictive and utility-based fingermark quality metric.

%% file: main_03_methods.tex
\section{Quality-aware fixed-length representation}
\label{sec:recognition}
The recognition stage trains the AFID encoder serving as the shared backbone for various downstream tasks, while producing a quality-aware fixed-length embedding used for identification. We propose an identity encoder that is invariant to rotation, translation, and scale within an operational range, requires no capture metadata at inference, and is trained on public data alone. 

\subsection{Architecture}
We use a convolutional neural network (CNN) as the main backbone of the approach, specifically the ConvNeXt-T v2 variant~\cite{liu2022convnext,woo2023convnextv2}, which combines beneficial properties of both convolutional and transformer-based architectures. While ViTs have become popular in recent years, particularly as foundation models, CNNs retain advantages in data-constrained training regimes. They incorporate spatial inductive bias by design, removing the need for positional embeddings or large pretraining datasets to learn this assumption. Their multi-scale feature maps are also directly compatible with U-Net-style decoders for downstream dense prediction tasks such as segmentation or minutiae extraction.

Let $E$ denote the ConvNeXt encoder, which maps an input image $\mathbf{x}$ to a feature map. The fixed-length identity embedding $\mathbf{y} = R(E(\mathbf{x})) \in \mathbb{R}^{768}$ is produced by the recognition head $R$, which applies GeM pooling to the final-stage features (of shape $768 \times 24 \times 24$) followed by batch normalization. In the downstream stages the encoder is frozen, denoted $\bar{E}$, and the feature extraction and quality assessment modules operate on its multi-stage feature map. The encoder contains approximately 28M trainable parameters. 

\subsection{Learning objective}
%Our goal is to train a recognition encoder that produces embeddings that are both discriminative and quality-aware. 
Our loss consists of two terms, reflecting the dual goal of identity discrimination and quality-awareness. The first, and the primary driver of identity discrimination, is the MagFace loss~\cite{meng2021magface}, which extends the ArcFace angular-margin objective~\cite{deng2019arcface} by making the margin and an auxiliary regularization term depend on the embedding magnitude $\|\mathbf{y}\|$:
\begin{equation}
\mathcal{L}_{\text{MF}} = \frac{1}{|\mathcal{B}_{\text{id}}|} \sum_{i \in \mathcal{B}_{\text{id}}} \Big[ \mathcal{L}_{\text{ang}}\!\left(\mathbf{y}_i; m(\|\mathbf{y}_i\|)\right) + \lambda_g \, g(\|\mathbf{y}_i\|) \Big],
\label{eq:magface}
\end{equation}
where $\mathcal{B}_{\text{id}}$ is the set of identity-labeled samples in the batch and $\mathbf{y}_i$ is the embedding of $i$-th sample. $\mathcal{L}_{\text{ang}}$ is the ArcFace angular-margin classification loss, $m(\cdot)$ is the magnitude-dependent margin replacing ArcFace's fixed margin, and $g(\cdot)$ is a regularizer that encourages large norms for easy samples and smaller norms for hard ones. The full formulation is given in~\cite{meng2021magface}. As a consequence of this design, $\|\mathbf{y}\|$ serves as a built-in quality signal, which we exploit in Section~\ref{sec:quality_norm}.

While face recognition typically assumes preprocessed inputs containing a center-aligned face, the same assumption is not realistic for fingermarks. An image recovered from a crime scene may contain a heavily smudged impression with no visible ridges or other background structures unrelated to the ridge content. Recognition training with angular margin losses requires a labeled identity for each image, so such no-identity samples are never observed during training. The resulting embeddings fall outside the distribution seen during training and may produce norms that no longer correlate with sample quality. 

To address this, we extend training to include no-identity background samples through an additional regularization term that pulls their embedding norms toward a fixed low-norm target:

\begin{equation}
\mathcal{L}_{\text{BG}} = \frac{1}{|\mathcal{B}_{\text{BG}}|} \sum_{i \in \mathcal{B}_{\text{BG}}} \left(\|\mathbf{y}_i\| - \tau\right)^2,
\label{eq:bg_reg}
\end{equation}

where $\mathcal{B}_{\text{BG}}$ represents the set of background samples in the current batch, $\|\mathbf{y}_i\|$ is the embedding norm of the $i$-th sample, and $\tau$ is the target norm for background samples. We set $\tau = 5$, below the lower bound of the MagFace norm range ($l_a=10$), which creates a clear gap between background samples and the lowest-quality labeled samples.

The background images themselves are drawn from three sources: blank images of uniform grayscale, manually selected fingermark images from public datasets containing no visible ridge content, and synthetically generated close-up images of common everyday objects and surfaces, generated using a commercial image generation model\footnote{Gemini 3.1 Flash Image model (Nano Banana 2)}. To avoid interfering with the recognition objective at the start, the background regularizer is activated at epoch 10 and its weight ramped linearly to $\lambda_{\text{BG}} = 0.01$ over the following 10 epochs, with background samples inserted into each batch with probability 0.05.

The final recognition loss combines the MagFace objective with the epoch-dependent background regularizer:
\begin{equation}
\mathcal{L}_{\text{recognition}} = \mathcal{L}_{\text{MF}} + \lambda_{\text{BG}}(t) \cdot \mathcal{L}_{\text{BG}}.
\label{eq:total_loss}
\end{equation}

\subsection{Input canonicalization and augmentation}
\label{sec:canonicalization}
Most recognition systems transform each input into a canonical form before extracting the representation. In face recognition, for example, the face is cropped and aligned beforehand. This is not realistic for fingermarks, where images are degraded, noisy, and cluttered with background. Rather than constructing a canonical input prior to extracting the embedding, we keep preprocessing minimal and let the model learn the variability it must tolerate directly from the data. To achieve this we make use of heavy data augmentation during training.

\textbf{Inference preprocessing.} Each image is scaled so its longer side matches the input size of 768 pixels while preserving aspect ratio, then zero-padded along the shorter side to a $768 \times 768$ square, and normalized to zero mean and unit variance with friction-ridge-specific channel statistics ($\mu=0.709$, $\sigma=0.256$). These are the only operations applied at inference; no resolution metadata, auxiliary signals, or alignment is required.

\textbf{Training augmentation.} We apply aggressive data augmentation during training. The most distinctive component is scale canonicalization: rather than assuming a known pixel density (PPI) at inference, we expose the model to a range of physical areas during training. The image physical area is computed from its dimensions and resolution, $A_{\text{mm}} = (\max(W_{\text{px}}, H_{\text{px}}) \cdot \tfrac{25.4}{\text{PPI}})^2$, a target area is sampled uniformly from $[1, 25]\,\text{cm}^2$, and the image is rescaled accordingly. This range spans small partial fingermarks up to full nail-to-nail rolled impressions. The rescaled image is randomly translated within the padded canvas and rotated uniformly in range $[-180^\circ, 180^\circ]$. Slap and rolled fingerprints are additionally occluded at the center, to discourage overfitting to the central region, or at the periphery, to simulate partial impressions. Photometric augmentation is applied per pixel: random intensity inversion (enforcing invariance to ridge polarity across development methods and sensors), Gaussian blur, brightness and contrast jitter. The only assumption imposed at inference is thus that the image is a friction ridge impression of roughly $1$--$25\,\text{cm}^2$, which any reasonably cropped impression satisfies by default.

\subsection{Training procedure}
The backbone is optimized with AdamW~\cite{loshchilov2019decoupled} using two parameter groups: a base learning rate of $2 \times 10^{-4}$ for the encoder and a higher learning rate of $4 \times 10^{-3}$ for the MagFace classification head. The schedule combines a linear warm-up over the first five epochs followed by cosine annealing to zero. Training is performed on a single NVIDIA H100 GPU with a batch size of 48. The model is trained for up to 60 epochs, and the epoch with the best identification performance on the validation set is selected for the final model. After training, only the encoder and the recognition head are retained and the MagFace classification head is discarded.

\section{Feature Extraction}
\label{sec:features}
During the second stage, we attach task-specific decoders, AFID-X, to the frozen encoder $\bar{E}$ to predict three feature representations widely used in friction ridge processing: a foreground segmentation mask separating ridge content from background, a minutiae probability map encoding minutiae locations and orientations, and a dense local ridge orientation field. These form the basic feature set used in both automated pipelines and manual examination by dactyloscopic experts.

\subsection{Decoder architecture}
\label{sec:decoder_arch}

Each task-specific head consists of a U-Net-style upsampling decoder~\cite{ronneberger2015unet} that takes as input the multi-scale feature maps of the frozen backbone encoder $\bar{E}$ and predicts an output of size $768 \times 768$. The decoder mirrors the four-stage hierarchy of the ConvNeXt backbone with four corresponding upsampling stages. Each stage involves bilinear upsampling, channel-wise concatenation with the matching encoder feature map via a skip connection, and two convolutional blocks with LayerNorm and GELU activations. A final $1 \times 1$ convolution maps the resulting features to the task-specific number of output channels, followed by additional $2 \times$ upsampling to restore the full input resolution. %Throughout the decoder, upsampling is performed by bilinear interpolation followed by a standard convolution rather than by transposed (strided) convolutions. In our experiments, this consistently produced smoother predictions and avoided any checkerboard artifacts. 
The same decoder architecture is used for all three prediction heads, only the number of output channels differs: one for segmentation, twelve for minutiae detection, and two for the local ridge orientation field. %The choice of output channels for each head is motivated in Section~\ref{sec:heads}. The total parameter count of a single decoder is approximately [TBD]\,M, which is roughly an order of magnitude smaller than the encoder.

\begin{figure}[t]
  \centering
  \includegraphics[width=\columnwidth]{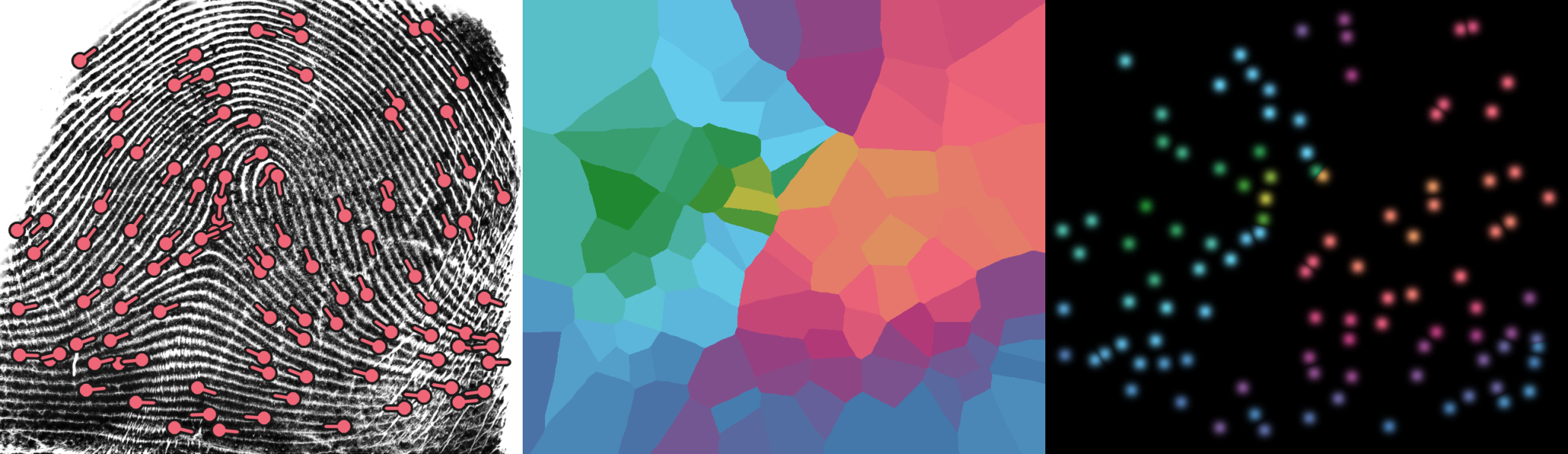}
  \caption{Learning local ridge orientation. For every training image with annotated minutiae points (left), we create a Voronoi diagram where each pixel is seeded with the modulo-$180^\circ$ orientation of the nearest minutia (center). During training, the local orientation decoder is supervised only at minutiae locations, where the orientation confidence is high (right). Angles are encoded with different colors for visualization purposes.}
  \label{fig:local_field_target}
\end{figure}

\subsection{Prediction heads and supervision}
\label{sec:heads}

\textbf{Foreground segmentation.} The segmentation head produces a single-channel output of size $768 \times 768$ corresponding to a binary mask separating the developed ridge region from the background. Supervision is given by manually annotated masks. The head is trained with a standard binary cross-entropy loss applied per pixel:
\begin{equation}
\mathcal{L}_{\text{seg}} = \text{BCE}(\sigma(\hat{\mathbf{S}}), \mathbf{S}),
\label{eq:loss_seg}
\end{equation}
where $\hat{\mathbf{S}} \in \mathbb{R}^{1 \times 768 \times 768}$ is the head's logit output, $\sigma$ is the sigmoid function, and $\mathbf{S} \in \{0,1\}^{1 \times 768 \times 768}$ is the ground-truth mask.

\textbf{Minutiae detection.} The minutiae head produces a $12 \times 768 \times 768$ output that encodes minutiae as a multi-channel heatmap, following~\cite{cao2020endtoend}. A minutia at $(x, y)$ with orientation $\theta$ is encoded by a spatial Gaussian centered at $(x, y)$ with standard deviation $\sigma = 0.2\,\text{mm}$, with each of the twelve channels spanning $30^\circ$ and $\theta$ interpolated across adjacent channels. We do not predict minutia type (endings vs bifurcation), as type information is often ambiguous and rarely used in operational automated matching pipelines. The head is trained with binary cross-entropy applied independently per channel:
\begin{equation}
\mathcal{L}_{\text{min}} = \frac{1}{12} \sum_{c=1}^{12} \text{BCE}(\sigma(\hat{\mathbf{M}}_c), \mathbf{M}_c),
\label{eq:loss_min}
\end{equation}
where $\hat{\mathbf{M}}_c$ and $\mathbf{M}_c$ are the predicted logits and ground-truth heatmap for the $c$-th channel.

\textbf{Local ridge orientation.} The orientation head produces a dense two-channel field $\hat{\mathbf{O}} \in \mathbb{R}^{2 \times 768 \times 768}$ encoding the local ridge orientation at each pixel as a unit vector $(\cos 2\theta, \sin 2\theta)$. The factor of two accounts for the $180^\circ$ symmetry of ridge orientation, yielding a continuous and periodic angle encoding.

Reliable dense ground-truth orientation fields are not available at scale for fingermarks, since their construction requires either manual annotation across the entire image or an existing reliable orientation estimator. We instead learn to predict a dense orientation field directly from sparse but high-confidence minutiae orientations. 

We first construct a dense but approximate target field: each pixel is assigned the modulo-$180^\circ$ orientation of its nearest annotated minutia, producing a Voronoi partition seeded at the minutiae locations. This target is accurate only near the minutiae themselves, where the local ridge orientation matches the minutia orientation, and becomes progressively unreliable with distance. We therefore weight the loss by a spatial confidence map $\mathbf{W} \in [0,1]^{1 \times 768 \times 768}$ that encodes the minutiae locations as Gaussians with $\sigma = 0.2\,\text{mm}$, concentrating the gradient signal on high-confidence pixels near the annotated minutiae. The orientation decoder is trained with a masked mean-squared error between its raw output $\hat{\mathbf{O}}$ and the unit-vector target $\mathbf{O}$:
\begin{equation}
\mathcal{L}_{\text{ori}} = \frac{1}{\sum_{i,j} \mathbf{W}_{i,j}} \sum_{i,j} \mathbf{W}_{i,j} \, \bigl\| \hat{\mathbf{O}}_{i,j} - \mathbf{O}_{i,j} \bigr\|^2 ,
\label{eq:loss_ori}
\end{equation}
where $\mathbf{O}_{i,j}$ is the target unit vector at pixel $(i,j)$. Despite this sparse supervision, the decoder learns to produce smooth and consistent orientation predictions across the entire image, including in regions far from any annotated minutia. The intermediate representations are shown in Figure~\ref{fig:local_field_target}. To the best of our knowledge, this is the first approach that proposes learning dense ridge orientation from sparse minutiae-derived angles.

\subsection{Training procedure}
\label{sec:feat_training}
The three decoders are attached to the frozen backbone $\bar{E}$ and share no parameters. They are trained jointly to minimize a weighted sum of the per-task losses:
\begin{equation}
\mathcal{L}_{\text{feat}} = \lambda_{\text{seg}} \mathcal{L}_{\text{seg}} + \lambda_{\text{min}} \mathcal{L}_{\text{min}} + \lambda_{\text{ori}} \mathcal{L}_{\text{ori}},
\label{eq:loss_feat}
\end{equation}
with $\lambda_{\text{seg}} = 0.25$, $\lambda_{\text{min}} = 5.0$, $\lambda_{\text{ori}} = 0.1$ balancing the dynamic ranges of the individual losses. Since the heads are independent, each is optimized with its own AdamW optimizer at a base learning rate of $3 \times 10^{-4}$, reduced on plateau, with best-validation checkpointing over up to 150 epochs. Input processing follows the recognition stage, adding random flipping, with supervision targets transformed alongside the image to preserve spatial alignment. All three heads are supervised on the same $5{,}050$ NIST~SD~302 images: minutiae locations and orientations from the NIST-provided templates, and foreground segmentation masks produced as part of this work and released with the framework.

\section{Quality Assessment}
\label{sec:quality}
The third stage of the framework, AFID-Q, is the quality assessment module. It produces a quality $Q \in [0, 100]$ for each input image, calibrated to the operational range specified by ISO/IEC~29794-1~\cite{iso29794-1:2024}. AFID-Q derives quality from two complementary signals, both computed on the frozen recognition encoder $\bar{E}$: an unsupervised signal $Q_n$ obtained directly from the MagFace embedding norm, and a supervised signal $Q_e$ produced by a small regressor trained on expert quality annotations. The two are combined into the final score $Q$ by a fusion step described in Section~\ref{sec:quality_fusion}.

\subsection{Quality from the embedding norm}
\label{sec:quality_norm}

The norm-based quality signal is a direct consequence of the recognition training stage. The MagFace loss (Eq.~\ref{eq:magface}) pushes the embeddings of recognizable samples away from the origin, so the norm $\|\mathbf{y}\|$ of an identity embedding correlates with the utility of its input image. The background regularizer (Eq.~\ref{eq:bg_reg}) extends this behavior to inputs that would otherwise fall outside the training distribution, pulling the norms of background and content-free samples toward the origin. The norm therefore varies monotonically with sample quality across the operational range, from background and degenerate inputs at the low end to clean, fully developed impressions at the high end. We take the embedding norm directly as the unsupervised quality signal, $Q_n(\mathbf{x}) = \|\mathbf{y}\|$.

\begin{figure}[t]
  \centering
  \includegraphics[width=\columnwidth]{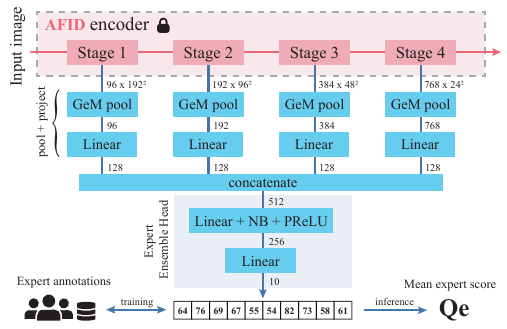}
  \caption{Expert quality assessment regressor. The decoder, supervised by expert-annotated quality values, receives features from all four stages of the AFID encoder. The features are pooled, projected and concatenated into a 512-d vector, which is then passed through a MLP that predicts an ensemble of 10 quality scores. To get the final quality value $Q_e$ we take the average of the predicted scores.}
  \label{fig:qe_head}
\end{figure}

\subsection{Quality from expert supervision}
\label{sec:quality_expert}

For fingermarks, the norm-based signal is weakened by the limited coverage of fingermark data in public training corpora. We therefore complement it with a supervised regressor that predicts expert quality scores, injecting experience from operational casework that the embedding norm alone does not capture.

\textbf{Supervision data.} The regressor is trained on the JRC annotation campaign~\cite{haraksim2023annotation}, in which $1{,}000$ NIST~SD~301 and SD~302 images were independently scored by ten certified examiners spanning eight EU member states, Australia, and Europol. Each image has quality annotations in range $[1, 10]$ from all ten examiners, which we rescale to $[0, 100]$. We retain the full ten-dimensional annotation vector per image rather than aggregating to a single score. This preserves the variability of expert opinion for the model to learn directly.

\textbf{Regressor architecture.} Unlike the recognition head, which pools the final stage into a single identity embedding, the expert regressor reads from all four stages of the frozen encoder $\bar{E}$, giving it access to the finer ridge detail present at the shallower levels. Each stage is reduced by an independent GeM pooling layer and projected to a shared $128$-dimensional width before the four are concatenated. Equalizing the widths gives every stage equal representation in the concatenated descriptor, so the head weights the scales by usefulness instead of their native channel count. A small MLP ($512 \rightarrow 256 \rightarrow 10$) then maps the descriptor to ten expert scores in range $[0, 1]$. The expert quality regressor is shown in Figure~\ref{fig:qe_head}. During inference, the final expert score $Q_e$ is produced by averaging all 10 predicted expert scores. 

\textbf{Training objective.} The regressor is trained with a mean squared error against the ten-dimensional expert annotation vector:
\begin{equation}
\mathcal{L}_{\text{exp}} = \frac{1}{10} \sum_{k=1}^{10} \left( \hat{q}^{(k)} - q^{(k)} \right)^2,
\label{eq:loss_exp}
\end{equation}
where $\hat{q}^{(k)}$ and $q^{(k)}$ are the predicted and annotated scores of the $k$-th expert. Treating each expert as a separate target encourages the regressor to capture the disagreement among examiners instead of only predicting an averaged consensus.

\subsection{Combining the signals}
\label{sec:quality_fusion}
The two quality signals are complementary: the norm-based score $Q_n$ is the stronger predictor for rolled and plain fingerprints, while the expert-supervised score $Q_e$ is the stronger predictor for fingermarks. A quality metric for the full friction ridge input space therefore benefits from fusing the two:
\begin{equation}
Q = Q_n \cdot Q_e .
\label{eq:fusion}
\end{equation}
The product imposes a useful inductive bias: a high final score requires both signals to agree, and either can veto the other when it is low. This approximates routing each input to its stronger predictor, but operates on the two scores alone and requires no knowledge of capture conditions at inference. Among the fusion rules we considered, including a weighted sum, minimum, and harmonic mean, the product gave the best ordering of samples by recognition utility.

\subsection{Training procedure}
\label{sec:quality_training}
The expert regressor is the only component of AFID-Q that requires training. It is trained on the frozen backbone $\bar{E}$ by minimizing the per-expert loss of Eq.~\ref{eq:loss_exp}, using AdamW at a learning rate of $3 \times 10^{-4}$ on a train/validation split of the 1,000 annotated images, and we retain the checkpoint with the lowest validation loss. Because quality depends on the amount of visible ridge content, augmentations that crop or discard ridge area would alter the target and are excluded. The regressor sees only content-preserving transformations: full-range rotation, horizontal flips, zoom-out-only scaling, small translations, intensity inversion, and mild brightness and contrast jitter.

%% file: main_04_experiments.tex
\begin{table}[t]
\centering
\caption{Datasets used in this work. Reported counts reflect the samples retained after cleanup, including the removal of mislabeled or corrupted images and those lacking the annotations required for our pipeline, and may therefore differ from the official dataset sizes.}
\label{tab:datasets}
\begin{threeparttable}
\begin{tabular}{lrrr}
\toprule
Dataset & Fingers & Fingerprints\tnote{1} & Fingermarks\tnote{1} \\
\midrule
\multicolumn{4}{l}{\textbf{Training pool}} \\
LFIW~\cite{liu2024lfiw}                 &      600 &           2{,}394  &  1{,}800 \\
IIIT-D MOLF~\cite{sankaran2015multisensor}                 &  1{,}000 &          11{,}934  &  4{,}400 (+15{,}958) \\
%Neurotechnology\tnote{2} &       77 &              928   &  -- \\
NIST~SD300~\cite{sd300}          &  8{,}880 &          17{,}659  &  -- (+35{,}318) \\
NIST~SD301~\cite{sd301}          &      240 &           2{,}762  &  -- (+3{,}908) \\
NIST~SD302~\cite{fiumara2018sd302}          &  1{,}800 & 22{,}832 &  4{,}207 (+28{,}800) \\

\cmidrule(lr){1-4}
Total                            & 12{,}520 & 57{,}581 &  10{,}407 (+83{,}984) \\
\midrule
\multicolumn{4}{l}{\textbf{Evaluation pool}} \\
FVC 2000~\cite{maio2002fvc2000}                  &  300\tnote{2}  &          2{,}400  & -- \\
FVC 2002~\cite{maio2002fvc2002}                  &  300\tnote{2}  &          2{,}400  & -- \\
FVC 2004~\cite{maio2004fvc2004}                   &  300\tnote{2} &          2{,}400 & -- \\
NIST~SD27~\cite{sd27}            &  2{,}040 &           2{,}040  &      258 \\
NIST~SD302~\cite{fiumara2018sd302}          &      200\tnote{3} &           2{,}566  &      455 \\
NIST~SD303~\cite{sd303}          & 160 & 160 &  6400 \\
Anguli synthetic             & -- & -- (+10{,}000) &  -- \\
\cmidrule(lr){1-4}
Total                            &  3{,}560 &          14{,}992 (+10{,}000)  &      6782 \\
\bottomrule
\end{tabular}
\begin{tablenotes}
\footnotesize
\item[1] Numbers in parentheses denote synthetically generated samples derived from the same person identities. 
\item[2] Only DB1a, DB2a and DB3a were used for evaluation per the official FVC evaluation protocol. 
\item[3] Last 200 fingers of the SD302 dataset were used for evaluation with no overlap with the training set identities. 
\end{tablenotes}
\end{threeparttable}
\end{table}

\begin{table*}[!t]
\centering
\caption{Verification performance on the FVC benchmark databases. Reported as EER and TAR @ 0.1\% FAR following the FVC protocol. All values are percentages. Best result per column is in \textbf{bold} and second best is \underline{underlined}.}
\label{tab:fvc_verification}
\begin{threeparttable}
\setlength{\tabcolsep}{3pt}
\footnotesize
\begin{tabular}{l cc cc cc cc cc cc cc cc cc}
\toprule
& \multicolumn{6}{c}{FVC 2000} & \multicolumn{6}{c}{FVC 2002} & \multicolumn{6}{c}{FVC 2004} \\
\cmidrule(lr){2-7} \cmidrule(lr){8-13} \cmidrule(lr){14-19}
& \multicolumn{2}{c}{DB1A} & \multicolumn{2}{c}{DB2A} & \multicolumn{2}{c}{DB3A}
& \multicolumn{2}{c}{DB1A} & \multicolumn{2}{c}{DB2A} & \multicolumn{2}{c}{DB3A}
& \multicolumn{2}{c}{DB1A} & \multicolumn{2}{c}{DB2A} & \multicolumn{2}{c}{DB3A} \\
\cmidrule(lr){2-3} \cmidrule(lr){4-5} \cmidrule(lr){6-7}
\cmidrule(lr){8-9} \cmidrule(lr){10-11} \cmidrule(lr){12-13}
\cmidrule(lr){14-15} \cmidrule(lr){16-17} \cmidrule(lr){18-19}
Matcher & EER & TAR & EER & TAR & EER & TAR
        & EER & TAR & EER & TAR & EER & TAR
        & EER & TAR & EER & TAR & EER & TAR \\
\midrule
VeriFinger 2025.1
& \textbf{0.07} & \textbf{99.96}
& \textbf{0.03} & \textbf{100.00}
& 0.40 & \textbf{99.39}
& \underline{0.25} & 99.64
& 0.39 & 99.57
& \textbf{0.39} & \textbf{99.50}
& \underline{0.28} & \underline{98.75}
& \textbf{0.39} & \underline{98.57}
& \textbf{0.10} & \underline{99.89} \\
AFR-Net\tnote{a}~\cite{grosz2024afrnet}
& -- & --   & -- & --   & -- & --
& -- & \underline{99.86} & -- & \textbf{99.96} & -- & \underline{98.43}
& -- & \textbf{100.00} & -- & \textbf{99.36} & -- & \textbf{100.00} \\
FDD~\cite{pan2024fdd}
& 0.61 & 99.29
& 0.18 & 99.79
& 0.54 & \underline{98.25}
& 0.28 & 99.68
 & 0.36 & 99.50
& 1.68 & 95.75
& 0.32 & 99.50
& 0.75 & 98.43
& 0.32 & 99.11 \\
FLARE~\cite{pan2026flare}
& 0.50 & 99.32
& 0.10 & \underline{99.89}
& \underline{0.29} & 99.21
& \textbf{0.08} & \textbf{99.93}
& \underline{0.22} & 99.71
& 1.18 & 96.61
& \textbf{0.22} & 99.68
& \underline{0.54} & 98.54
& \underline{0.25} & 99.54 \\
\textbf{AFID }
& \underline{0.22} & \underline{99.75}
& \underline{0.04} & \textbf{100.00}
& \textbf{0.28} & \textbf{99.39}
& 0.39 & 99.18
& \textbf{0.11} & \underline{99.82}
& \underline{0.86} & 97.46
& 0.32 & 99.18
& 0.61 & 98.36
& 0.50 & 99.18 \\
\bottomrule
\end{tabular}
\begin{tablenotes}
\footnotesize
\item [a] AFR-Net values as reported by Grosz and Jain~\cite{grosz2024afrnet}. EER and FVC 2000 results are not reported by the authors.
\end{tablenotes}
\end{threeparttable}
\end{table*}

\section{Experiments}
\label{sec:experiments}

In this section we present the results of the proposed framework. We first describe the experimental environment, then show the recognition performance of the AFID encoder, demonstrate the feature extraction capabilities of the AFID-X decoders and finally, present the predictive power of the proposed AFID-Q quality assessment metric.

\subsection{Datasets}
Our model is trained on publicly available datasets only\footnote{This includes unrestricted datasets and some licensed datasets accessible upon request to the dataset owners.}. Table~\ref{tab:datasets} shows the train and evaluation splits, with no identity overlap between them so that recognition performance is measured on unseen fingers. The images span a range of sources and sensors, from high-resolution optical scans and contactless finger photos to distorted operational fingermarks.

The training pool comprises approximately $68{,}000$ real images across $12{,}520$ fingers, from LFIW~\cite{liu2024lfiw}, IIIT-D MOLF~\cite{sankaran2015multisensor}, and NIST~SD~300--302~\cite{sd300,sd301,fiumara2018sd302}. For evaluation we use the FVC benchmarks~\cite{maio2002fvc2000,maio2002fvc2002,maio2004fvc2004} and three fingermark datasets. NIST~SD~27~\cite{sd27} has been discontinued but is retained for comparison with prior work. NIST~SD~302 was recently revised with additional annotated fingermarks; we reserve the last 200 fingers ($455$ fingermarks) strictly for evaluation. NIST~SD~303~\cite{sd303}, released in 2026, has to our knowledge not yet been used as a fingermark recognition benchmark. To evaluate identification performance, we expand each gallery with $10{,}000$ Anguli~\cite{anguli_fingerprint} synthetic distractors. We additionally generate synthetic training images by compositing labeled impressions onto cluttered backgrounds, detailed in the supplementary material.

Supervision for the feature extraction decoders come from NIST minutiae templates (locations and orientations, $5{,}050$ images), while we annotated the segmentation masks for the same images. The JRC annotation campaign expert quality scores~\cite{haraksim2023annotation} ($1{,}000$ SD~301/302 images) supervise the quality regressor.

\subsection{Metrics}
For fingerprint-to-fingerprint verification we follow the FVC protocol and report the equal error rate (EER) and true accept rate (TAR) at $0.1\%$ false accept rate (FAR); for fingermark-to-fingerprint verification we additionally report the stricter TAR at $0.01\%$ FAR. Identification is reported through Cumulative Match Characteristic (CMC) curves and rank-$n$ identification rates. For the AFID-X decoders, minutiae extraction is assessed by precision, recall, and F1 against the NIST-provided minutiae. For segmentation performance, we report intersection-over-union (IoU), missing-detection, and false-detection rates against manually annotated masks. To assess the ridge orientation estimation, we compute root-mean-square deviation (RMSD) between predicted and ground-truth block orientations. AFID-Q quality assessment is evaluated with error-versus-discard characteristic (EDC) curves, summarized by the normalized partial area under the curve (nAUC).

\begin{table}[t]
\centering
\caption{Fingermark-to-fingerprint verification performance against an extended gallery. Reported as TAR @ FAR $= 0.1\%$ and FAR $= 0.01\%$. All values are percentages. Best result per column in \textbf{bold}.}
\label{tab:latent_verification}
\begin{threeparttable}
\setlength{\tabcolsep}{4pt}
\footnotesize
\begin{tabular}{l cc cc cc}
\toprule
& \multicolumn{2}{c}{SD~27} & \multicolumn{2}{c}{SD~302} & \multicolumn{2}{c}{SD~303} \\
\cmidrule(lr){2-3} \cmidrule(lr){4-5} \cmidrule(lr){6-7}
& \multicolumn{6}{c}{TAR @ FAR} \\
%\cmidrule(lr){2-7}
Matcher & 0.1\% & 0.01\% & 0.1\% & 0.01\% & 0.1\% & 0.01\% \\
\midrule
VeriFinger 2025.1
& 67.21 & \textbf{56.56}
& 50.77 & 42.42
& \textbf{67.25} & 60.17 \\
FDD~\cite{pan2024fdd}
& 49.59 & 38.52
& 21.32 & 14.07
& 60.06 & 51.02 \\
FLARE~\cite{pan2026flare}
& 62.70 & 49.59
& 23.52 & 15.82
& 61.41 & 52.08 \\
\textbf{AFID }
& \textbf{70.49} & \textbf{56.56}
& \textbf{54.51} & \textbf{45.05}
& 65.50 & \textbf{60.97} \\
\bottomrule
\end{tabular}
\end{threeparttable}
\end{table}

\subsection{Verification performance}
\label{sec:exp_verification}
We first report 1:1 verification performance, following established protocols~\cite{engelsma2021deepprint,grosz2024afrnet,pan2026flare}. We compare AFID against the commercial VeriFinger 2025.1 and two publicly released models, FDD~\cite{pan2024fdd} and FLARE~\cite{pan2026flare}. AFR-Net~\cite{grosz2024afrnet} is unreleased, so we quote its published numbers where the protocol is comparable. Table~\ref{tab:fvc_verification} reports fingerprint-to-fingerprint matching on the FVC benchmarks, and Table~\ref{tab:latent_verification} fingermark-to-fingerprint matching on NIST~SD~27, SD~302, and SD~303.

On FVC, all methods perform at or near saturation and the differences are small. Averaged across the nine databases, TAR@0.1\% is $99.47\%$ for VeriFinger, $99.16\%$ for FLARE, $99.15\%$ for AFID, and $98.81\%$ for FDD. VeriFinger holds a small lead, AFID and FLARE are effectively tied among the open methods, and AFR-Net leads on several of the databases it reports. The methods separate most on the hardest database, FVC~2002~DB3A, where partial impressions from a small capacitive sensor drop TAR for all methods and AFID ranks highest among the open ones. AFID trails only on the two FVC~2004 databases with deliberately introduced skin distortion. Overall, AFID is competitive with established methods on clean fingerprint verification, matching the best open matchers and trailing the commercial reference only slightly.

Fingermark-to-fingerprint verification (Table~\ref{tab:latent_verification}) is substantially harder, and here AFID is strongest overall. It achieves the highest TAR on SD~27 and SD~302 at both operating points and the highest TAR@0.01\% on SD~303, and it surpasses even VeriFinger on SD~27 and SD~302. The margin over the other open methods is largest on SD~302, where AFID reaches a TAR@0.1\% of $54.51\%$, more than double FLARE's $23.52\%$ and FDD's $21.32\%$.

\begin{figure*}[t]
  \centering
  \includegraphics[width=\textwidth]{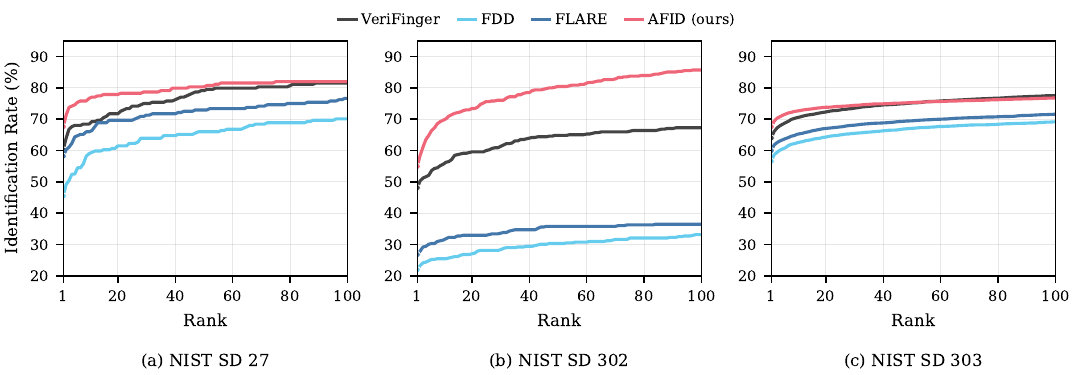}
  \caption{Cumulative Match Characteristic (CMC) curves for fingermark identification against the extended 10K gallery on (a) NIST SD~27, (b) NIST SD~302, and (c) NIST SD~303. AFID consistently leads across the operationally relevant rank range.}
  \label{fig:cmc_curves}
\end{figure*}

\begin{table}[t]
\centering
\caption{Fingermark identification performance against an extended 10K fingerprint gallery. Reported as rank-$k$ identification rate (\%). Best result per column in \textbf{bold}.}
\label{tab:identification}
\begin{threeparttable}
\setlength{\tabcolsep}{3pt}
\footnotesize
\begin{tabular}{l ccc ccc ccc}
\toprule
& \multicolumn{3}{c}{SD~27} & \multicolumn{3}{c}{SD~302} & \multicolumn{3}{c}{SD~303} \\
\cmidrule(lr){2-4} \cmidrule(lr){5-7} \cmidrule(lr){8-10}
Matcher & R1 & R10 & R20 & R1 & R10 & R20 & R1 & R10 & R20 \\
\midrule
VeriFinger 2025.1
& 60.2 & 68.4 & 71.7
& 48.1 & 55.6 & 59.6
& 64.0 & 70.5 & 72.3 \\
FDD~\cite{pan2024fdd}
& 45.5 & 59.0 & 61.5
& 22.0 & 25.5 & 27.0
& 56.6 & 62.5 & 64.3 \\
FLARE~\cite{pan2026flare}
& 58.2 & 66.0 & 69.7
& 26.8 & 31.4 & 33.0
& 60.0 & 65.2 & 67.0 \\
\textbf{AFID}
& \textbf{67.6} & \textbf{76.6} & \textbf{77.9}
& \textbf{54.9} & \textbf{69.7} & \textbf{73.4}
& \textbf{67.6} & \textbf{72.6} & \textbf{73.8} \\
\bottomrule
\end{tabular}
\end{threeparttable}
\end{table}

\subsection{Identification performance}
\label{sec:exp_identification}
We report closed-set identification (1:$N$) performance, in which each fingermark probe is matched against a gallery of reference fingerprints. As is common in related literature~\cite{cao2020endtoend,grosz2024afrnet}, we augment the gallery of each dataset (SD~27, SD~302, and SD~303) with $10{,}000$ distractor impressions, extending it beyond the original dataset gallery size, which better reflects the search space of an operational system. Table~\ref{tab:identification} reports rank-1, rank-10, and rank-20 identification rates for AFID, FDD, FLARE, and VeriFinger 2025.1 as an operational reference. Figure~\ref{fig:cmc_curves} shows the full Cumulative Match Characteristic (CMC) curves up to rank 100.

AFID achieves the highest identification rate on all three datasets, by a margin that varies with dataset difficulty. The separation is largest on SD~302, where  marks are frequently partial, superimposed, and arbitrarily rotated. AFID reaches a rank-1 rate of $54.9\%$, ahead of VeriFinger ($48.1\%$) and far ahead of FLARE ($26.8\%$) and FDD ($22.0\%$). FDD and FLARE trail AFID only moderately on SD~27 and SD~303 but collapse to roughly half its rank-1 rate on SD~302, a degradation concentrated on the dataset with the most varied orientations that points to an orientation sensitivity AFID does not share. The lead narrows on SD~303, composed of upright partial marks whose smudging and degradations involve less geometric variation, so all methods perform more evenly.

Direct comparison with AFR-Net is not possible. It was evaluated against a $100{,}000$-image gallery and was never released publicly. As an indirect reference, AFR-Net reports a rank-1 rate of $53.10\%$, on par with or slightly below VeriFinger v12.3 under its protocol, whereas AFID exceeds the newer VeriFinger 2025.1 on SD~27 under ours. With VeriFinger as a shared reference point this is a favorable indication for AFID, though the differing galleries and VeriFinger versions preclude a direct comparison.

The performance gap on SD~302 points to a broader difference in how AFID handles orientation. Most recent recognition pipelines align the input explicitly, estimating a canonical orientation and rotating the impression upright before extracting the representation~\cite{grosz2024afrnet,pan2024fdd,pan2026flare}. This assumes a single upright orientation can be recovered from the input, which holds for complete impressions, where singular points and global ridge flow can determine the finger orientation, but breaks down for small partial fingermarks that lack these cues. AFID instead learns an orientation-invariant representation through the rotation augmentation described in Section~\ref{sec:recognition}, encoding a superposition of orientations, rather than committing to a single canonical one. We analyze this invariance in detail in section~\ref{sup:rotation} of the supplementary material, showing that AFID embeddings remain near-invariant across the full $360^\circ$ and that the same invariance holds at the level of local descriptors as well.

\begin{table}[t]
\centering
\caption{Ablation of AFID design choices, reported as rank-1 identification rate (\%) against the extended 10K gallery. Each row removes or replaces a single component of the full model. Cell shading indicates change relative to the full model. Best result per column in \textbf{bold}.}
\label{tab:ablation}
\begin{threeparttable}
\setlength{\tabcolsep}{5pt}
\footnotesize
\begin{tabular*}{\linewidth}{@{\extracolsep{\fill}} l cccc}
\toprule
Model & SD27 & SD302 & SD303 & Mean \\
\midrule
\textbf{AFID} (\textit{MagFace, $768^2$}) & 67.6 & 55.2 & 67.6 & \textbf{63.5} \\
\midrule
\multicolumn{5}{@{}l}{\itshape Augmentation} \\
\quad no scale
  & \cellcolor{red!20}59.0
  & \cellcolor{red!5}53.9
  & \cellcolor{green!8}68.2
  & \cellcolor{red!12}60.4 \\
\quad no rotation
  & \cellcolor{red!35}53.3
  & \cellcolor{red!60}29.9
  & \cellcolor{green!10}\textbf{68.9}
  & \cellcolor{red!40}50.7 \\
\quad no inversion
  & \cellcolor{red!12}61.5
  & \cellcolor{red!3}54.1
  & \cellcolor{red!3}67.1
  & \cellcolor{red!10}60.9 \\
\midrule
\multicolumn{5}{@{}l}{\itshape Data and regularization} \\
\quad no synthetic data
  & \cellcolor{red!55}36.9
  & \cellcolor{red!12}50.8
  & \cellcolor{red!8}64.6
  & \cellcolor{red!40}50.7 \\
\quad no background reg.
  & \cellcolor{red!2}67.2
  & \cellcolor{green!3}\textbf{55.4}
  & \cellcolor{green!1}67.7
  & \cellcolor{red!1}63.4 \\
\midrule
\multicolumn{5}{@{}l}{\itshape Input image size} \\
\quad $512^2$
  & \cellcolor{red!22}57.8
  & \cellcolor{red!6}53.4
  & \cellcolor{red!6}66.0
  & \cellcolor{red!15}59.1 \\
\quad $1024^2$
  & \cellcolor{green!2}\textbf{68.0}
  & \cellcolor{red!12}50.8
  & \cellcolor{red!7}65.0
  & \cellcolor{red!8}61.3 \\
\midrule
\multicolumn{5}{@{}l}{\itshape Loss} \\
\quad ArcFace~\cite{deng2019arcface}
  & \cellcolor{red!15}63.1
  & \cellcolor{red!10}52.1
  & \cellcolor{green!1}67.8
  & \cellcolor{red!9}61.0 \\
\quad AdaFace~\cite{kim2022adaface}
  & \cellcolor{red!18}61.0
  & \cellcolor{red!16}49.7
  & \cellcolor{red!11}63.4
  & \cellcolor{red!18}58.1 \\
\bottomrule
\end{tabular*}
\end{threeparttable}
\end{table}

\subsection{Identity encoder ablation}
\label{sec:ablation}
To isolate the contribution of each design choice, we retrain the AFID identity encoder with individual components removed or replaced and report rank-1 identification on the extended gallery (Table~\ref{tab:ablation}). %The mean across the three benchmarks summarizes the overall effect, while the per-dataset columns reveal where each choice makes the biggest impact.

The augmentation components account for the largest share of performance. Removing rotation augmentation matters most. Mean rank-1 falls from $63.5\%$ to $50.7\%$, and the drop is concentrated almost entirely on SD~302, which collapses from $54.9\%$ to $29.9\%$ while SD~303 is essentially unaffected. The rotation-invariant representation is designed to handle the rotated marks that dominate SD~302, and removing the augmentation reproduces the orientation sensitivity seen in the alignment-based baselines. Scale augmentation and intensity inversion contribute less, with the scale effect most visible on SD~27, where the marks are not already tightly cropped. SD~303 is largely insensitive to the augmentation and scale choices due to being composed of upright partial slap marks with limited geometric variation.

Synthetic training data is essential, particularly for SD~27, which contains the most varied backgrounds. The synthetic marks are generated with substantial background clutter, so that the model learns invariance to background-specific features. Removing synthetic data lowers mean rank-1 to $50.7\%$ and SD~27 rank-1 from $67.6\%$ to $36.9\%$.

Input resolution is optimal at $768^2$: $512^2$ costs over four mean points as ridge detail is lost, while $1024^2$ offers no gain. The background regularizer has little effect on recognition, consistent with its role of stabilizing norms on out-of-distribution inputs, rather than aiding identification. Among losses, MagFace outperforms both ArcFace and AdaFace. While AdaFace provides a quality signal through its embedding norm, it suppresses hard, low-quality samples as presumed noise, which is counterproductive for fingermarks since those are the operationally relevant samples. MagFace thus gives the best recognition in addition to the quality signal it provides.

%To isolate each design choice, we retrain the encoder with individual components removed or replaced and report rank-1 identification (Table~\ref{tab:ablation}). Rotation augmentation matters most: removing it drops mean rank-1 from $63.5\%$ to $50.7\%$, almost entirely on SD~302 ($54.9\% \rightarrow 29.9\%$) while the upright SD~303 is unaffected, reproducing the orientation sensitivity of the alignment-based baselines on exactly the rotated marks it is meant to handle. Scale augmentation and intensity inversion contribute a few points each, the scale effect most visible on the less tightly cropped SD~27. Synthetic data is likewise essential, particularly on SD~27 ($67.6\% \rightarrow 36.9\%$ without it), whose varied backgrounds the synthetic clutter teaches the model to ignore.

%Input resolution is optimal at $768^2$: $512^2$ costs over four mean points as ridge detail is lost, while $1024^2$ offers no gain. The background regularizer has little effect on identification, consistent with its role of stabilizing norms on out-of-distribution inputs rather than aiding recognition. Replacing MagFace with ArcFace leaves recognition essentially unchanged, confirming that MagFace is chosen not for a recognition advantage but for the quality signal its magnitude-aware margin provides at no cost to identification.

% Minutiae extraction on latents 
\begin{table}[t]
\centering
\caption{Minutiae extraction performance on NIST~SD27 \textbf{fingermark probes} against NIST-provided ground-truth templates. A predicted minutia is counted as correct when it falls within $D$ pixels and $O^\circ$ of a ground-truth minutia; we report two threshold settings used by Nguyen et al.~\cite{nguyen2018minutiaenet}. All values are percentages. Best result per column in \textbf{bold}.}
\label{tab:minutiae_extraction}
\begin{threeparttable}
\setlength{\tabcolsep}{6pt}
\footnotesize
\begin{tabular*}{\linewidth}{l @{\extracolsep{\fill}} ccc @{\extracolsep{\fill}} ccc}
\toprule
 & \multicolumn{3}{c}{$D = 12,\ O = 20^\circ$} & \multicolumn{3}{c}{$D = 16,\ O = 30^\circ$} \\
\cmidrule(lr){2-4} \cmidrule(lr){5-7}
Method & Precision & Recall & F1 & Precision & Recall & F1 \\
\midrule

FDD~\cite{pan2024fdd}\tnote{a}           & 51.5 & 47.5 & 49.4 & 53.9 & 49.7 & 51.7 \\
Verifinger 2025.1                        & 53.0 & 51.1 & 52.1 & 55.4 & 53.4 & 54.4 \\
FingerNet\tnote{b}~\cite{tang2017fingernet}       & 58.0 & 58.1 & 58.0 & 63.0 & 63.2 & 63.1 \\

MinutiaeNet-I\tnote{c}~\cite{nguyen2018minutiaenet} & 69.2 & 63.3 &  66.1 & 72.6 & 66.4 & 69.4  \\
MinutiaeNet-R\tnote{c}~\cite{nguyen2018minutiaenet} & \textbf{70.5} & \textbf{72.3} & \textbf{71.4} & \textbf{71.2} & \textbf{75.7} & \textbf{73.4} \\
\textbf{AFID-X }                    & 66.5  &  65.4   & 66.0  &  69.2  & 68.1  &  68.7 \\
\bottomrule
\end{tabular*}
\begin{tablenotes}
\footnotesize
\item [a] Extracted based on the minutiae probability map that FDD pipeline (version using voting pose estimation + PriorEnh enhancement) produces internally.
\item [b] Results of FingerNet are shown as reported by the authors.
\item [c] For MinutiaeNet, we show the results reported in~\cite{nguyen2018minutiaenet} (R) and the results achieved with the publicly released implementation on GitHub (I). 
\end{tablenotes}
\end{threeparttable}
\end{table}

% Feature extraction figure 
\begin{figure}[t]
\centering
\setlength{\tabcolsep}{2pt}
\begin{tabular}{cc}
\includegraphics[width=\columnwidth]{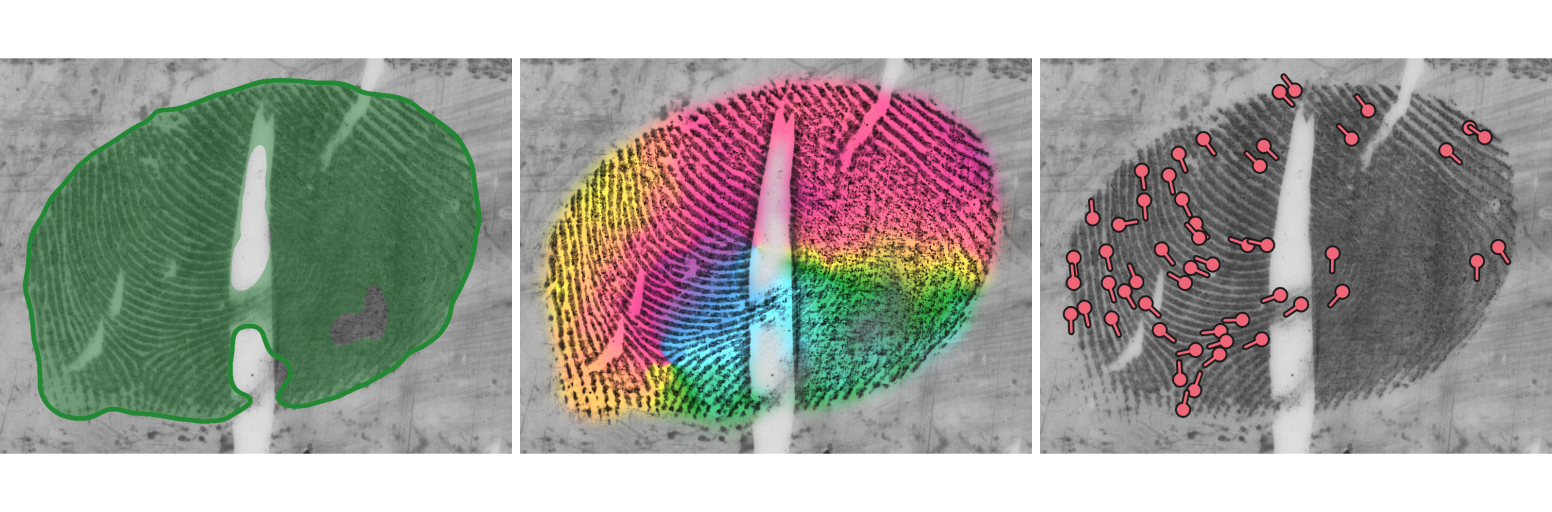} \\%$[2pt]

\includegraphics[width=\columnwidth]{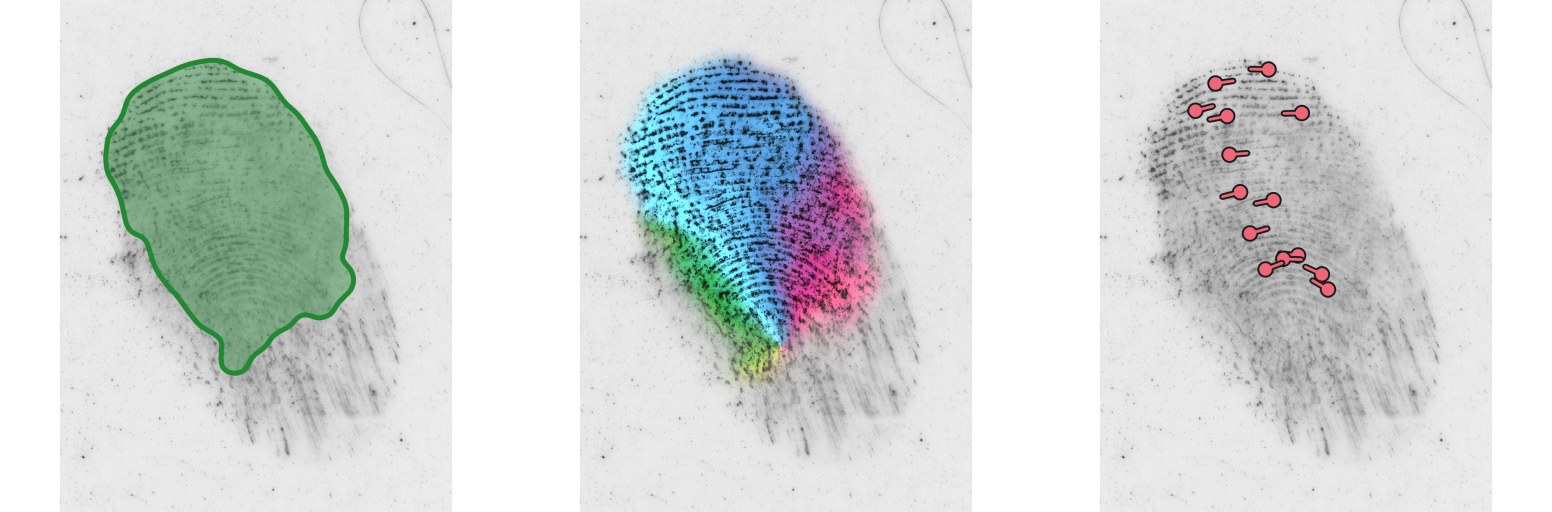}\\ %$&

\includegraphics[width=\columnwidth]{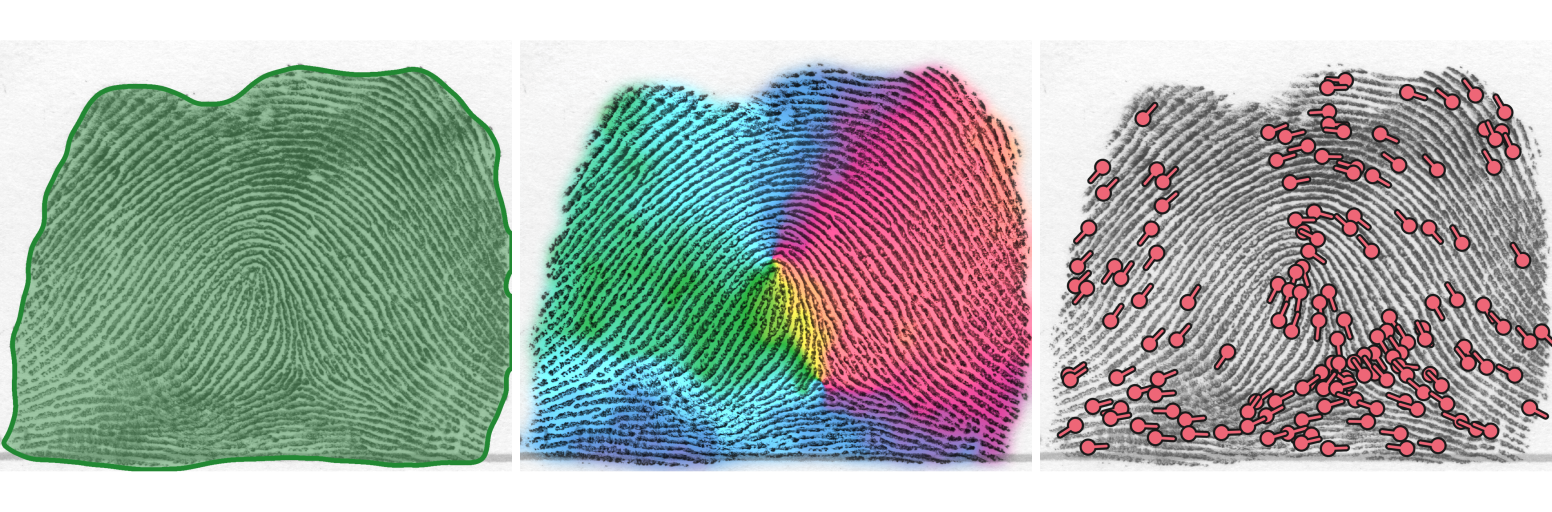} \\

\includegraphics[width=\columnwidth]{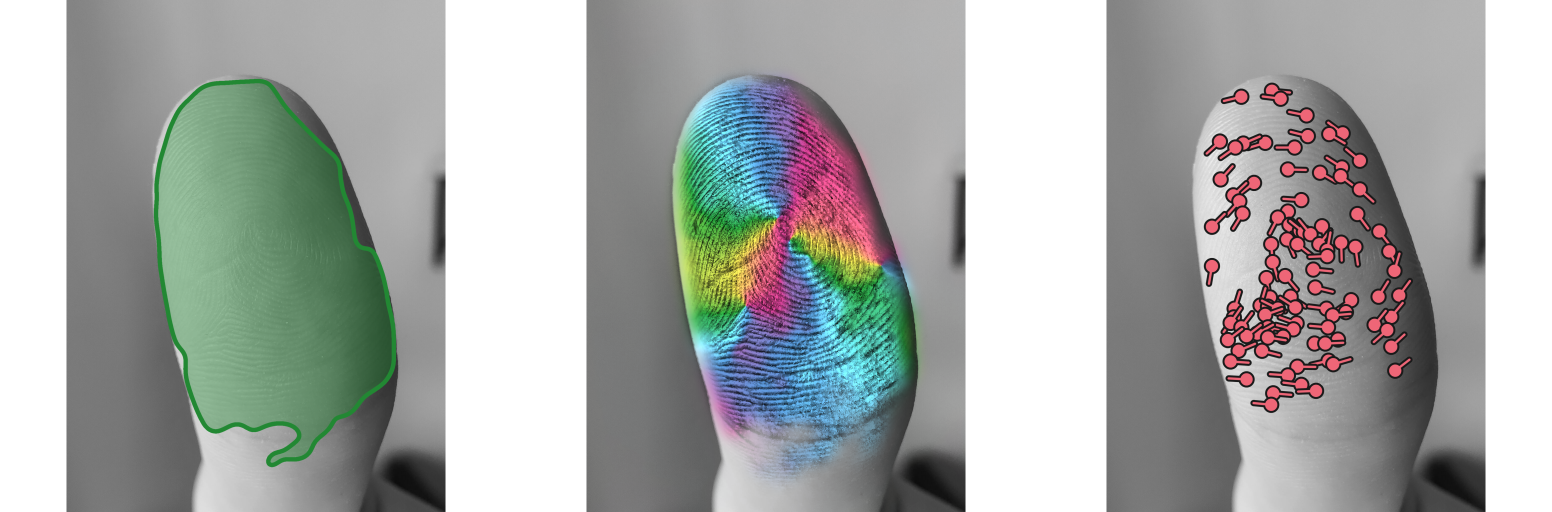}\\

\end{tabular}
\caption{%
    Predicted features for four impressions. Each triplet shows,
    left to right, the predicted \emph{segmentation mask}, \emph{local ridge orientation} and the detected \emph{minutiae points}. Ridge orientation is defined modulo $180^{\circ}$ and is encoded with a continuous color scheme:
    \orikey{0}{4477AA}~$0^{\circ}$ (horizontal),
    \orikey{45}{43A98F}~$45^{\circ}$,
    \orikey{90}{CCBA44}~$90^{\circ}$ (vertical) and
    \orikey{135}{CC4C77}~$135^{\circ}$,
    returning to \orikey{0}{4477AA}~$0^{\circ}$ at $180^{\circ}$. 
    %The feature extraction is demonstrated on various kinds of fingerprint impressions including noisy partials (top row), low contrast (second row), rolled and slap fingerprints (third row), and finger photos (last row). All features are computed on raw images without preprocessing
  }
\label{fig:grid}
\end{figure}

\subsection{Minutiae extraction}
\label{sec:exp_minutiae}
We evaluate the AFID-X minutiae decoder against the minutiae templates distributed with NIST~SD~27. A predicted minutia is counted as correct when its location and orientation fall within distance $D$ and angle $O$ of an unmatched ground-truth minutia, and we report the two threshold settings used by Nguyen et al.~\cite{nguyen2018minutiaenet}. We compare against FingerNet~\cite{tang2017fingernet}, MinutiaeNet~\cite{nguyen2018minutiaenet}, VeriFinger, and FDD~\cite{pan2024fdd}. For MinutiaeNet we report both the numbers published in the original paper (denoted R) and those obtained with the publicly released implementation (denoted I). For AFID-X and FDD we select a detection threshold by calibrating against the SD~302 minutiae annotations. Results on the SD~27 fingermark probes are reported in Table~\ref{tab:minutiae_extraction}; results on the rolled exemplars are provided in the supplementary material. Figure~\ref{fig:grid} shows qualitative feature extraction on impressions from several datasets.

On SD~27 fingermarks, AFID-X places third, behind both MinutiaeNet variants and ahead of FingerNet, VeriFinger, and FDD. At the looser threshold it reaches an F1 of $68.7\%$, within a few points of the reported MinutiaeNet result ($73.4\%$), despite the decoder operating on a frozen recognition backbone that receives no minutiae supervision. This indicates the AFID encoder learned local ridge structure during recognition-focused training. We also observe that VeriFinger produces a substantial number of spurious minutiae on the noisier fingermarks, which lowers its precision on this set. On the SD~27 rolled exemplars the ordering reverses: AFID-X reaches an F1 of $85.3\%$, the highest of all tested methods, ahead of VeriFinger and FDD. We report the full rolled-exemplar comparison, along with downstream matching performance using Bozorth3 and MCC matchers, in the supplementary material.

\subsection{Segmentation and Local Orientation}
\label{subsec:seg_orient}

We evaluate the two AFID-X decoders against the manual Tsinghua markup of NIST SD27~\cite{feng2012orientation,yang2014localized}, the de facto reference annotation for latent segmentation and orientation on this benchmark.

For segmentation, AFID-X achieves an IoU of $59.43\%$, with a missing-detection rate of $14.97\%$ and a false-detection rate of $6.53\%$. These figures should be read with the subjectivity of latent segmentation ground truth in mind: the reference masks cover only the primary friction-ridge impression, whereas AFID-X recovers \emph{all} friction-ridge regions present in the image, including secondary and partial impressions that the annotation deliberately omits. A large share of the apparent overlap gap is therefore attributable to this annotation convention, rather than to spurious background response, and the reported IoU understates the qualitative accuracy of the predicted masks. For context, the best pixel-wise result reported on the same benchmark is SegFinNet ($81.76\%$ IoU)~\cite{nguyen2018segfinnet}.

For local ridge orientation, AFID-X is competitive with the state of the art. Following the standard protocol, we report the root-mean-square deviation (RMSD, in degrees) between the predicted and manually marked block orientations over valid foreground blocks, partitioned into the Good, Bad and Ugly subsets of SD27~\cite{feng2012orientation}. AFID-X achieves an average RMSD of $\textbf{12.55}^{\circ}$ (Good $10.27^{\circ}$, Bad $13.20^{\circ}$, Ugly $14.73^{\circ}$). This
places AFID-X on par with the strongest fingermark-specific method reported on this annotation, RefNet~\cite{duan2021refnet} ($12.10^{\circ}$ average), and ahead of earlier estimators from Cao et al.~\cite{cao2015latent} ($13.51^{\circ}$), Yang et al.~\cite{yang2014localized} ($14.35^{\circ}$) and FingerNet~\cite{tang2017fingernet} ($17.82^{\circ}$)\footnote{As reported by~\cite{duan2021refnet}.}. We regard this as a notable result given that, unlike these methods, the AFID-X orientation decoder is never supervised with a dense orientation field, but instead trained purely from the sparse, high-confidence angle annotations carried by minutiae points, yet recovers a smooth and accurate field across the full impression.

\begin{figure*}[t]
  \centering
  \includegraphics[width=\textwidth]{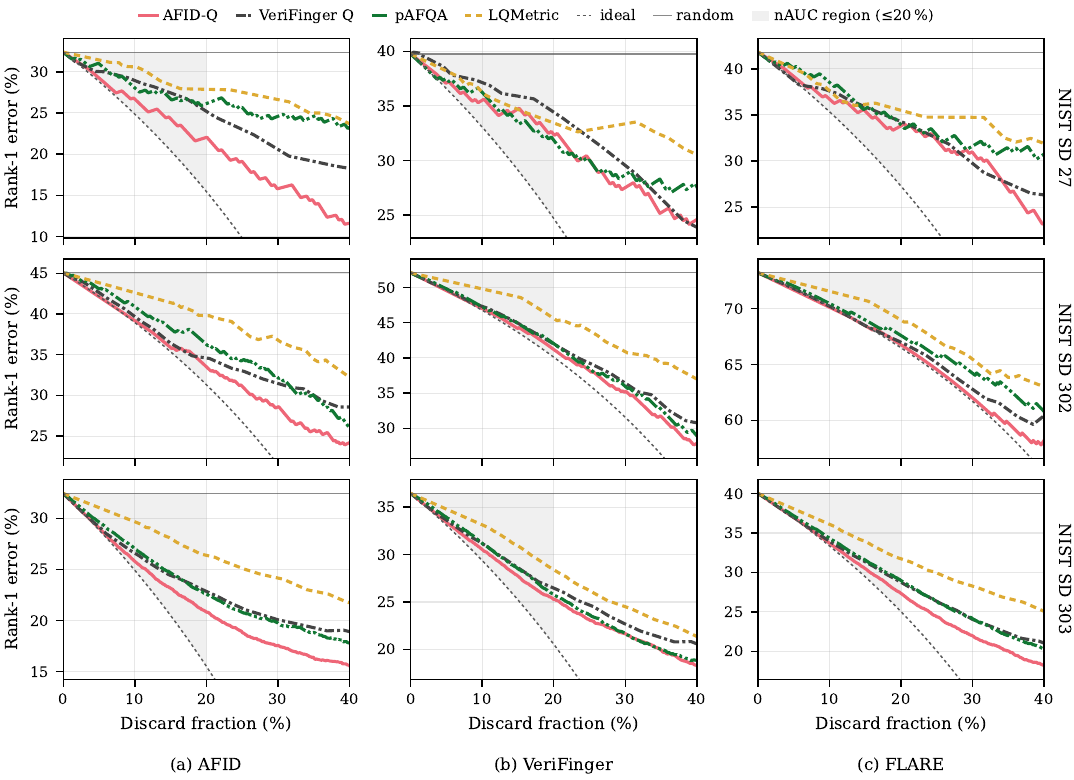}
  \caption{Rank-1 identification EDC. Probes are discarded worst-quality-first;
  the curve reports the rank-1 error over the probes that remain. Columns are
  matchers, rows are datasets. The dashed diagonal is the oracle ordering
  (every error discarded first) and the horizontal line is an uninformative
  ordering; the dotted vertical marks the 20\% discard limit over which the
  summary statistic of Table~\ref{tab:nauc-rank1} is integrated. Lower is
  better.}
  \label{fig:edc-ident}
\end{figure*}

\subsection{Quality assessment}
\label{sec:quality}

To evaluate how predictive the AFID-Q quality values are, we follow the error-versus-discard characteristic (EDC) defined in ISO/IEC~29794-1, but modify the setting slightly: instead of computing false non-match ratio (FNMR) at a fixed false match ratio (FMR) vs discard rate, which is more suitable for verification scenarios, we show rank-1 identification error ($1 - R1$). This is more suitable for the forensic context, where an AFIS returns a candidate list and no thresholds are defined. We evaluate on the three held-out fingermark datasets with the extended 10K gallery, repeating every experiment with three matchers (AFID, VeriFinger 2025.1, FLARE), since a quality metric must generalize across matchers rather than predict only its own. We compare AFID-Q against VeriFinger's quality value, pAFQA, and LQMetric, and report its two fused components, the embedding norm ($Q_n$) and expert-ensemble head ($Q_e$), as an ablation.

Figure~\ref{fig:edc-ident} shows the rank-1 EDC for all nine combinations, and Table~\ref{tab:nauc-rank1} summarizes it as the normalized partial AUC (nAUC) over the first $20\%$ of discards ($0$ = ideal, $1$ = uninformative). AFID-Q lies below every baseline over almost the entire discard range and obtains the lowest nAUC for all three matchers, leading in 8 of the 9 matcher/dataset combinations. On SD~27 with AFID, discarding the worst $20\%$ of marks lowers rank-1 error from $32.4\%$ to $22.0\%$ against an ideal floor of $15.5\%$, closing $61\%$ of the gap. The ablation in Table~\ref{tab:nauc-rank1} also confirms the fusion helps: AFID-Q improves on $Q_n$ in all nine scenarios and on $Q_e$ in seven, and $Q_e$ alone already outperforms every other existing method. Because $Q_n$ is derived from the AFID model, part of AFID-Q's advantage on the AFID matcher reflects their shared representation. Its improvement over the strongest baseline is indeed largest on AFID ($45\%$) and smallest on VeriFinger ($17\%$), but the advantage holds across all three matchers.

\begin{table}[t]
\centering
\caption{Rank-1 identification EDC, normalised pAUC over the first 20\% of discards ($0$~=~ideal discard curve, $1$~=~uninformative; lower is better). Each entry averages NIST~SD~27, SD~302, and SD~303. Best result per column in \textbf{bold}.}
\label{tab:nauc-rank1}
\begin{threeparttable}
\setlength{\tabcolsep}{6pt}
\footnotesize
\begin{tabular*}{\linewidth}{@{\extracolsep{\fill}} l r r r r}
\toprule
Quality Metric & AFID & VeriFinger & FLARE & Mean \\
\midrule
LQMetric & 0.648 & 0.526 & 0.432 & 0.535 \\
pAFQA & 0.399 & 0.278 & 0.253 & 0.310 \\
VeriFinger Q & 0.328 & 0.346 & 0.179 & 0.285 \\
\textbf{AFID-Q} & \textbf{0.182} & \textbf{0.231} & \textbf{0.132} & \textbf{0.182} \\
\midrule
\multicolumn{5}{l}{\itshape Ablation: individual components of AFID-Q} \\
$Q_n$ (embedding norm) & 0.287 & 0.300 & 0.189 & 0.259 \\
$Q_e$ (expert ensemble) & 0.204 & 0.243 & 0.157 & 0.202 \\
\bottomrule
\end{tabular*}
\end{threeparttable}
\end{table}

%% file: main_05_conclusion.tex
\section{Conclusion}
\label{sec:conclusion}
We presented AFID, a unified and fully open framework for friction ridge processing, built on a single recognition backbone trained on public data alone. Instead of extracting features to enable recognition, AFID inverts the conventional order, first learning a fixed-length identity representation and then deriving quality assessment and feature extraction from it on a shared frozen encoder. The representation sets a new state of the art in fixed-length fingermark recognition, leading identification across NIST~SD~27, SD~302, and SD~303 and surpassing the commercial VeriFinger matcher on fingermarks, with essentially no preprocessing required at inference time. Sitting on top of the frozen recognition encoder, AFID-Q predicts fingermark quality more accurately than every compared baseline across three matchers, and the AFID-X decoders recover segmentation, orientation, and minutiae points comparable to other methods, despite the recognition encoder receiving no feature-level supervision during training. %Although our evaluation focuses on fingermarks, the AFID framework generalizes across the full range of friction ridge imagery, including ink, optical, and capacitive impressions, slap and rolled prints, finger photos, and partial or distorted fingermarks.

Some limitations remain, and they motivate our future work. Segmentation is the weakest of the three extraction tasks: trained only on manually annotated masks, the decoder tends to detect all ridge structure in an image rather than isolating the central impression. The AFID-Q metric, while strongly predictive, also warrants further analysis across capture conditions and operating points, which we see as the path toward a validated open contribution to the ISO/IEC~29794-12 fingermark quality standard. By releasing the framework, the pretrained models, and the annotations produced in this work, we provide a reproducible foundation for the forensic community.

%% file: main_06_supplementary.tex
\clearpage
\onecolumn  % switch to single column for the title banner
\setcounter{page}{1}

\twocolumn[
  \begin{center}
    {\large\bfseries Supplementary Material\par}
    \vspace{1ex}
    %{\normalsize Tim Oblak, Rudolf Haraksim, Peter Peer\par}
    \vspace{2ex}
  \end{center}
]

\section{Synthetic Data Generation and Augmentation}
\label{sup:synthetic}
Alongside real impressions, we make extensive use of synthetically generated fingermarks, which expand the training distribution far beyond what the available real data covers. Each synthetic mark is generated from a clean rolled or plain fingerprint, its segmentation mask, and its annotated minutiae, composited onto a randomly sampled background surface. Because the source print's identity and annotations are carried through the pipeline, the resulting marks serve as valid labelled samples for both recognition and feature-extraction training. In total we produce approximately $84{,}000$ synthetic fingermarks.

\begin{figure}[t]
  \centering
  \includegraphics[width=\columnwidth]{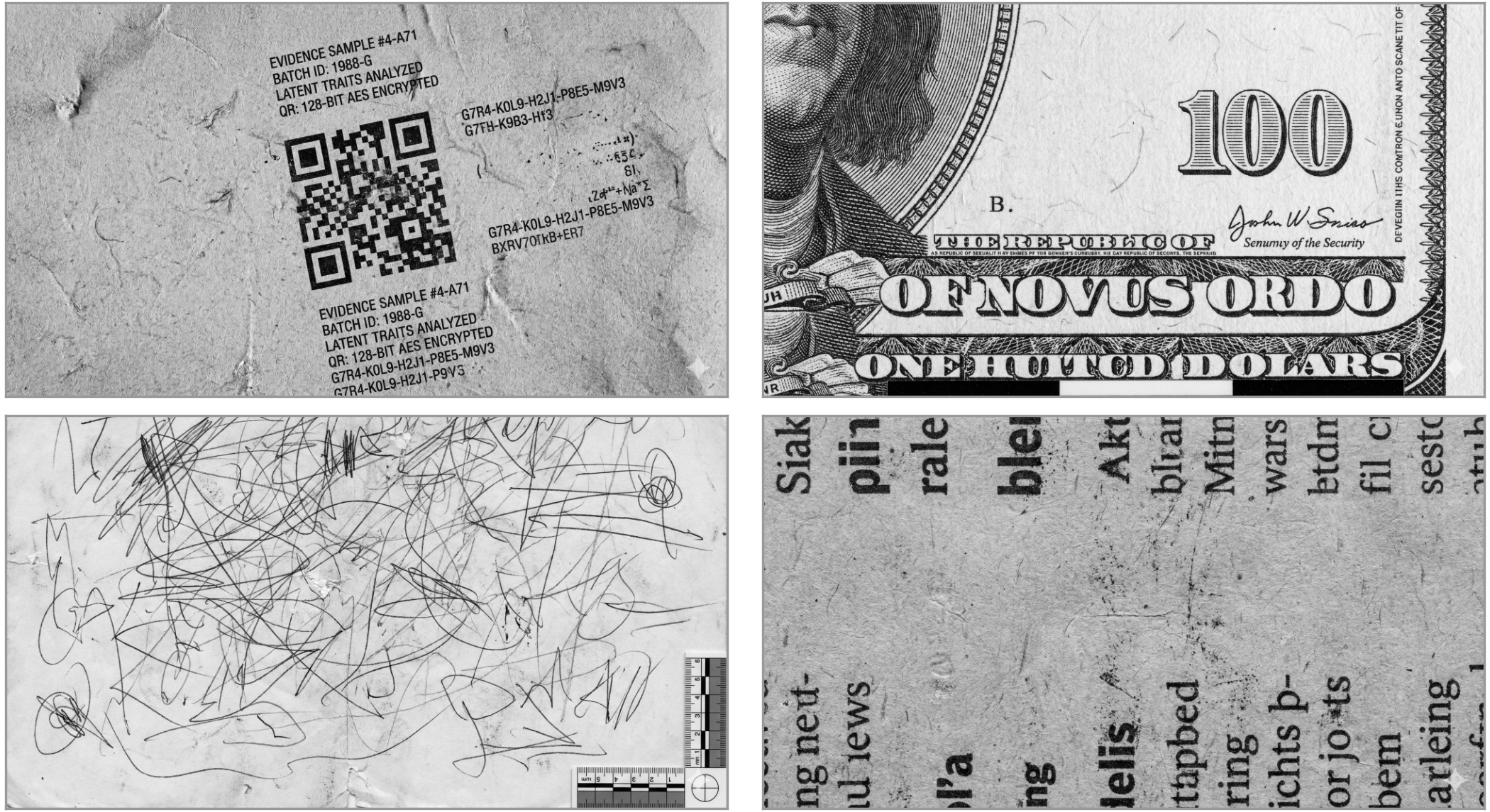}
  \caption{Examples of synthetic surface backgrounds used for background regularization during training, as well as to create synthetic fingermarks images.}
  \label{fig:bg_examples}
\end{figure}

\textbf{Background surfaces.} Realistic backgrounds are essential: operational fingermarks are recovered from an uncontrolled environment, and the background regularizer of Section~\ref{sec:recognition} additionally requires no-ridge surfaces resembling plausible deposition sites. We generate a collection of 50 surface photographs with a commercial image generation model\footnote{Gemini 3.1 Flash Image (Nano Banana 2).}, using a fixed prompt that varies only the surface:
\begin{quote}
\footnotesize
\textit{``Generate a forensic-style macro photograph of the surface of X. The camera angle is strictly top-down and perpendicular to the surface. Professional documentary lighting, neutral and even, to highlight micro-textures, scratches, and grain. High depth of field with the entire surface in sharp focus. No human hands, no visible fingerprints, and no artistic bokeh. Do not add a forensic scale. Realistic, unedited raw photo. The object fills the entire frame.''}
\end{quote}
The constraints on angle, lighting, focus, and the absence of hands or existing ridge content yield clean, evenly-lit textures suitable as deposition backgrounds. Examples are shown in Figure~\ref{fig:bg_examples}.

\begin{figure}[t]
  \centering
  \includegraphics[width=\columnwidth]{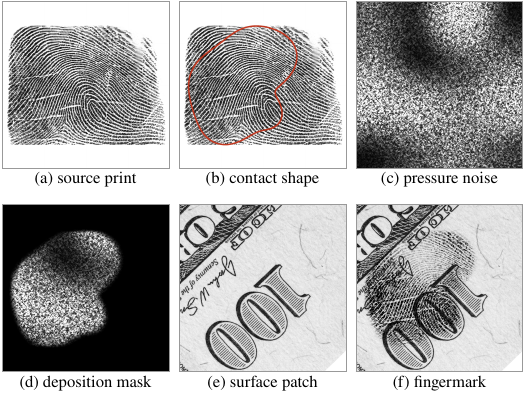}
  \caption{Synthetic fingermark generation. A clean rolled print (a) is masked by
a random-shape contact region (b) and a 2D perlin noise pressure
field (c), giving the deposition mask (d). The resulting residue is composited
onto a random surface patch (e) to yield the synthetic fingermark (f).
Segmentation and minutiae annotations propagate through the pipeline and are filtered based on the newly generated shape.}
  \label{fig:latent_pipeline}
\end{figure}

\textbf{Composition pipeline.} The source fingerprint is cropped to its segmentation mask, inverted so that ridges read as dark deposits on a light substrate, and placed on the output canvas under a random affine transform. Three effects then shape the deposit. A \emph{pressure map}, formed as the product of Perlin-like noise fields at multiple resolutions ($4$ to $256$ px), introduces realistic contact variation, with regions of strong contact fading into weak regions. A \emph{partial-contact mask} restricts the visible area to an organically perturbed ellipse, centred toward the finger core via the mask's distance transform. A randomly cropped and rotated \emph{background patch} provides the context. The final per-pixel opacity is the product of these masks with the eroded segmentation mask, and the inverted ridges are blended additively or subtractively depending on substrate brightness. The full pipeline is illustrated in Figure~\ref{fig:latent_pipeline}.

\textbf{Annotation propagation.} The affine transform applied to the image is applied analytically to the minutiae coordinates and orientations, transferring labels without interpolation error; minutiae falling outside the canvas or in regions where the visibility mask drops below a threshold are discarded. The segmentation mask is transformed likewise. Each synthetic sample is therefore an (image, mask, minutiae) triplet whose labels are exactly consistent with the composite, requiring no manual annotation.

\begin{figure*}[!t]
  \centering
  \includegraphics[width=\textwidth]{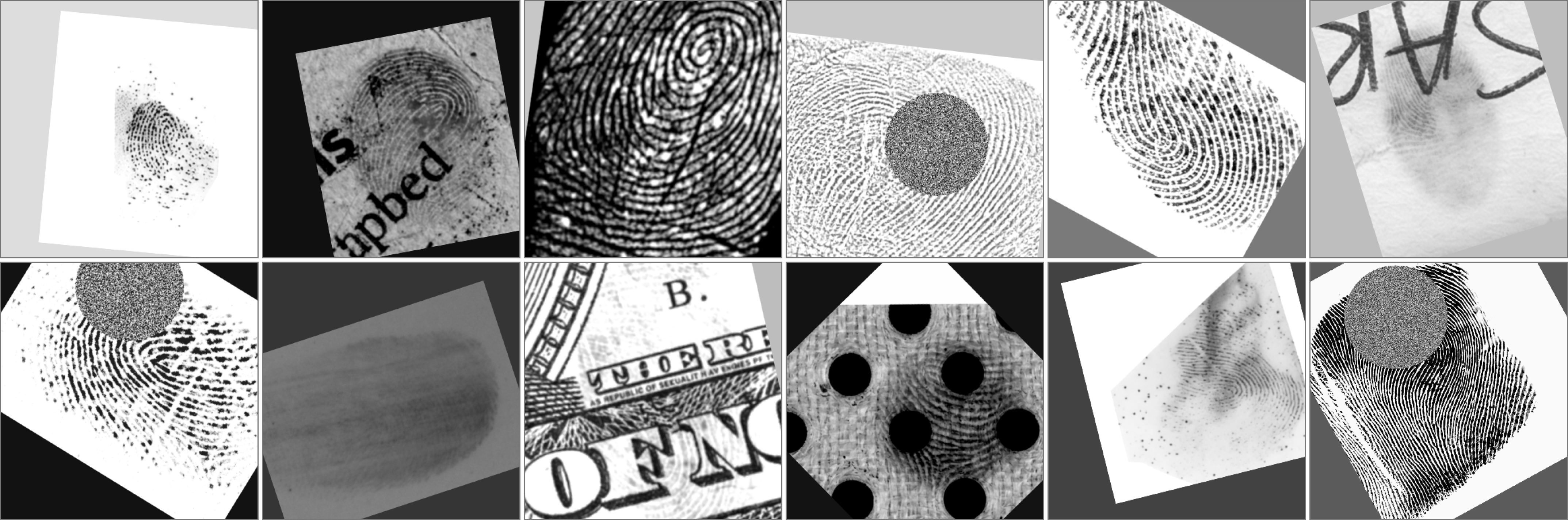}
  \caption{A random training batch after augmentation, illustrating the range of backgrounds, scales, rotations, occlusions, and photometric variation the encoder encounters.}
  \label{fig:batch_example}
\end{figure*}

\textbf{Training-time augmentation.} Like the real impressions, synthetic marks pass through the training augmentation of Section~\ref{sec:recognition}, adding area-based rescaling, full-range rotation, translation, occlusion, and photometric jitter. Figure~\ref{fig:batch_example} shows a representative augmented batch of real and synthetic friction ridge images, that can be encountered during training. Each of the presented images is labeled with it's own identity label.

\pagebreak

\section{Additional minutiae experiments}
\label{sup:rolled_minutiae}
We complement the minutiae extraction results on probe fingermarks in the main text with results on rolled gallery impressions. Table~\ref{tab:minutiae_rolled} reports minutiae extraction on the SD~27 rolled prints. Here the ordering of the methods changes: AFID-X achieves the highest F1 of all evaluated methods ($85.3\%$ at the looser detection threshold), ahead of both VeriFinger and FDD. The publicly released MinutiaeNet implementation performs poorly here, which we traced to common orientation errors on high-quality impressions, where a large fraction of its predicted minutiae are offset by $180^\circ$ from the ground truth. As the original paper does not report a rolled-exemplar benchmark, only the released implementation can be evaluated in this setting, so this result should be read as a property of that implementation rather than a property of the method as published. AFID-X's advantage over VeriFinger and FDD holds independently of MinutiaeNet.

Agreement with annotated minutiae measures how faithfully an extractor reproduces expert annotation, but not how useful the minutiae are for matching. We evaluate downstream utility by feeding each extractor's minutiae into two classical matchers, Bozorth3 and MCC~\cite{cappelli2010mcc}, and measuring the resulting identification rate (Table~\ref{tab:downstream}). Because ground-truth templates exist only for the $258$ SD~27 probes and their mated references, this experiment uses the native SD~27 gallery rather than the extended gallery of the main identification results, so the absolute rank-$k$ rates are not comparable to the CMC curves in the main text. The NIST ground-truth minutiae provide the performance ceiling for each matcher. Under both matchers, AFID-X minutiae yield the highest identification rate of any automated extractor by a wide margin, reaching a rank-1 rate of $58.1\%$ with MCC against $38.4\%$ for VeriFinger, and recovering roughly three quarters of the ground-truth ceiling. Although AFID-X ranks only third on agreement, the minutiae it extracts have high utility and are useful for downstream identification.

\newpage

% Minutiae extraction on rolled
\begin{table}[!h]
\centering
\caption{Minutiae extraction performance on NIST~SD27 \textbf{rolled exemplars} against NIST-provided ground-truth templates. A predicted minutia is counted as correct when it falls within $D$ pixels and $O^\circ$ of a ground-truth minutia; we report two threshold settings used by Nguyen et al.~\cite{nguyen2018minutiaenet}. All values are percentages. Best result per column in \textbf{bold}.}
\label{tab:minutiae_rolled}
\begin{threeparttable}
\setlength{\tabcolsep}{6pt}
\footnotesize
\begin{tabular*}{\linewidth}{l @{\extracolsep{\fill}} ccc @{\extracolsep{\fill}} ccc}
\toprule
 & \multicolumn{3}{c}{$D = 12,\ O = 20^\circ$} & \multicolumn{3}{c}{$D = 16,\ O = 30^\circ$} \\
\cmidrule(lr){2-4} \cmidrule(lr){5-7}
Method & Precision & Recall & F1 & Precision & Recall & F1 \\
\midrule
MinutiaeNet-I\tnote{c}~\cite{nguyen2018minutiaenet} & 41.7 & 37.9 & 39.7 & 46.8 & 42.5 & 44.6 \\
FDD~\cite{pan2024fdd}\tnote{a}           & 88.1 & 59.6 & 71.1 & 90.5 & 61.2 & 73.0 \\
Verifinger 2025.1                        & 67.6 & 78.8 & 72.8 & 69.7 & 81.2 & 75.0 \\
\textbf{AFID-X }                   & \textbf{83.4} & \textbf{87.0}  & \textbf{85.2}   & \textbf{85.0}  &  \textbf{88.7}  & \textbf{86.8} \\

\bottomrule
\end{tabular*}
\begin{tablenotes}
\footnotesize
\item [a] Extracted based on the minutiae probability map that FDD pipeline (voting pose estimation + PriorEnh enhancement) produces internally. 
\item [c] Results achieved with the publicly released MinutiaeNet implementation on GitHub; \cite{nguyen2018minutiaenet} does not report a rolled-print/reference-image benchmark.
\end{tablenotes}
\end{threeparttable}
\end{table}
% Downstream utility minutiae
\begin{table}[!h]
\centering
\caption{Downstream utility of extracted minutiae, matched with the open Bozorth3 and MCC matchers. We report SD~27 identification rate of 258 probes against the 258 rolled fingerprints. Ground-truth minutiae annotations are shown as an upper bound. Best result per column in \textbf{bold}.}
\label{tab:downstream}
\begin{threeparttable}
\setlength{\tabcolsep}{6pt}
\footnotesize
\begin{tabular*}{\linewidth}{@{\extracolsep{\fill}} l ccc ccc}
\toprule
& \multicolumn{3}{c}{Bozorth3} & \multicolumn{3}{c}{MCC} \\
\cmidrule(lr){2-4} \cmidrule(lr){5-7}
Method & R1 & R10 & R20 & R1 & R10 & R20 \\
\midrule
NIST GT\tnote{a}                                    & 53.9 & 61.6 & 65.5 & 74.8 & 83.3 & 86.0 \\
\midrule
MinutiaeNet-I\tnote{c}~\cite{nguyen2018minutiaenet} & 10.5 & 17.1 & 21.7 & 12.4 & 24.4 & 30.6 \\
Verifinger 2025.1                                   & 22.5 & 32.6 & 35.7 & 38.4 & 46.5 & 48.1 \\
FDD~\cite{pan2024fdd}\tnote{b}                      & 31.4 & 44.6 & 47.3 & 32.2 & 45.7 & 49.6 \\
\textbf{AFID-X}                              & \textbf{43.4} & \textbf{57.0} & \textbf{59.7} & \textbf{58.1} & \textbf{68.6} & \textbf{70.9} \\
\bottomrule
\end{tabular*}
\begin{tablenotes}
\footnotesize
\item [a] NIST-annotated ground-truth minutiae, shown as an upper bound.
\item [b] Extracted based on the minutiae probability map that FDD pipeline (voting pose estimation + PriorEnh enhancement) produces internally. Detection threshold selected based on best F1 score achieved.
\item [c] Results achieved with the publicly released MinutiaeNet implementation on GitHub.

\end{tablenotes}
\end{threeparttable}
\end{table}

\newpage
\section{Rotation invariance of AFID embeddings}
\label{sup:rotation}
\begin{figure}[!h]
  \centering
  \includegraphics[width=\columnwidth]{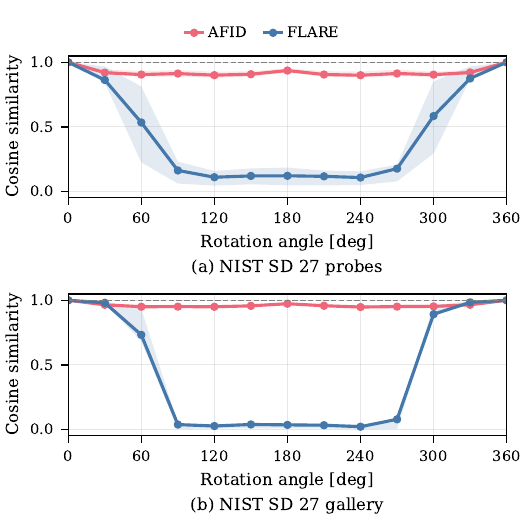}
  \caption{Rotational invariance on (a) 200 fingermark probes and (b) 200 rolled reference fingerprints. Each image is rotated in $30^\circ$ increments and its embedding compared against that of its unrotated version. The curve shows the mean and the shaded band the interquartile range.}
  \label{fig:rotation}
\end{figure}

Figure~\ref{fig:rotation} quantifies the rotation invariance discussed in the identification results. Each image is rotated in $30^\circ$ increments and its embedding compared, by cosine similarity, against the embedding of the unrotated version. AFID is nearly invariant across the full $360^\circ$ range, retaining a mean similarity above $0.93$ on fingermark probes and above $0.97$ on rolled gallery prints even at the worst-case $180^\circ$ offset. FLARE, in contrast, retains high similarity only within roughly $\pm 45^\circ$ of the original orientation before collapsing. This reflects the two designs directly, as AFID learns invariance through rotation augmentation while FLARE estimates and corrects a canonical pose that becomes underdetermined once the rotation exceeds the range its pose estimator was trained to handle.

\begin{figure}[]
\centering
\begin{subfigure}{\columnwidth}
    \includegraphics[width=\linewidth]{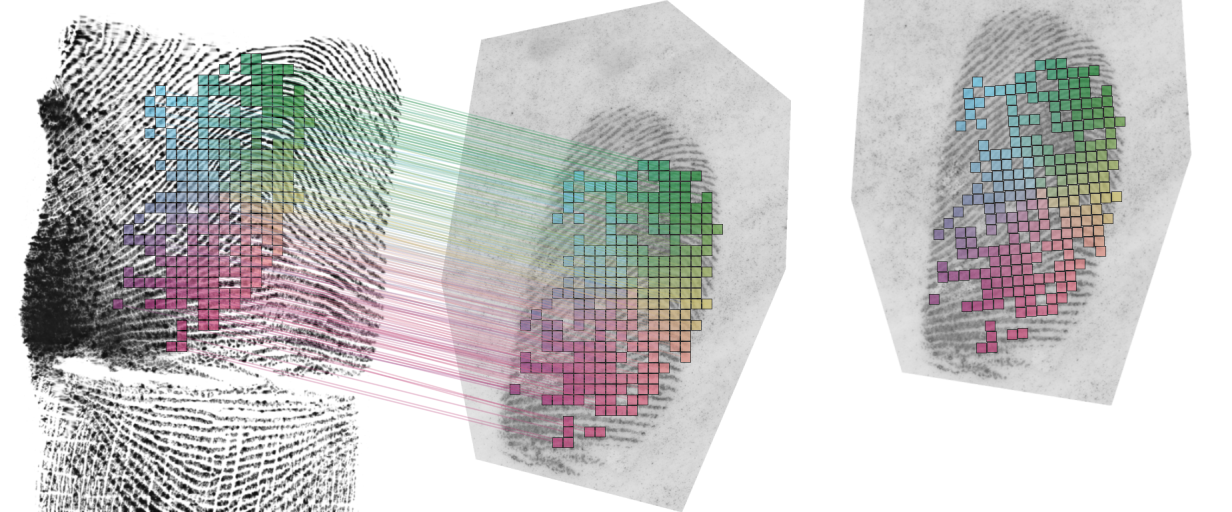}
    \caption{Cosine similarity: 0.761}
\end{subfigure}
\\[2pt]
\begin{subfigure}{\columnwidth}
    \includegraphics[width=\linewidth]{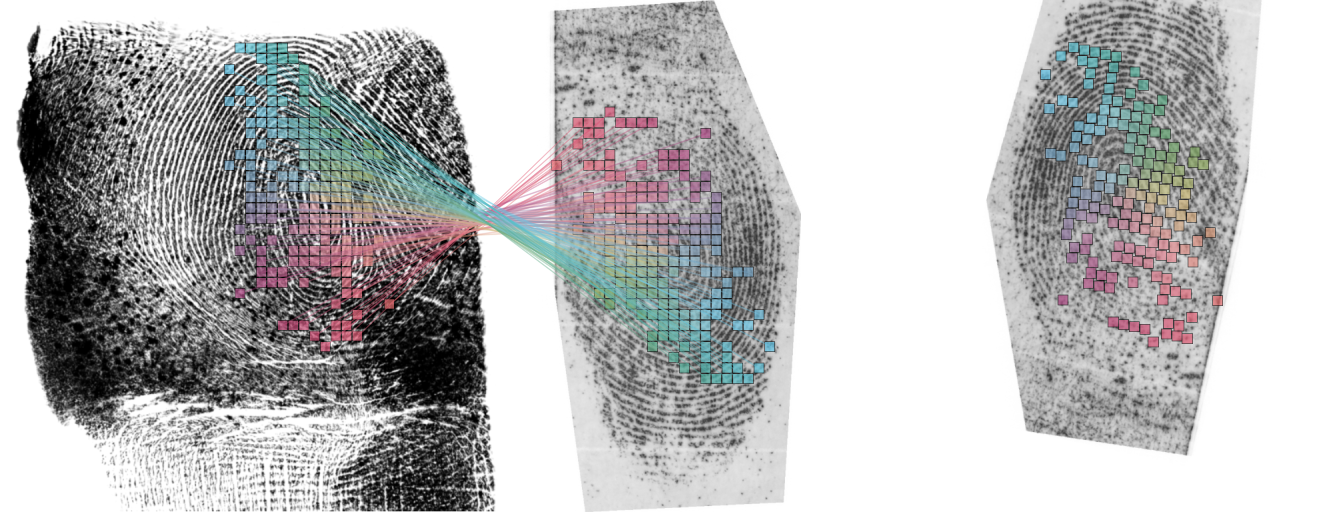}
    \caption{Cosine similarity: 0.706}
\end{subfigure}
\\[2pt]
\begin{subfigure}{\columnwidth}
    \includegraphics[width=\linewidth]{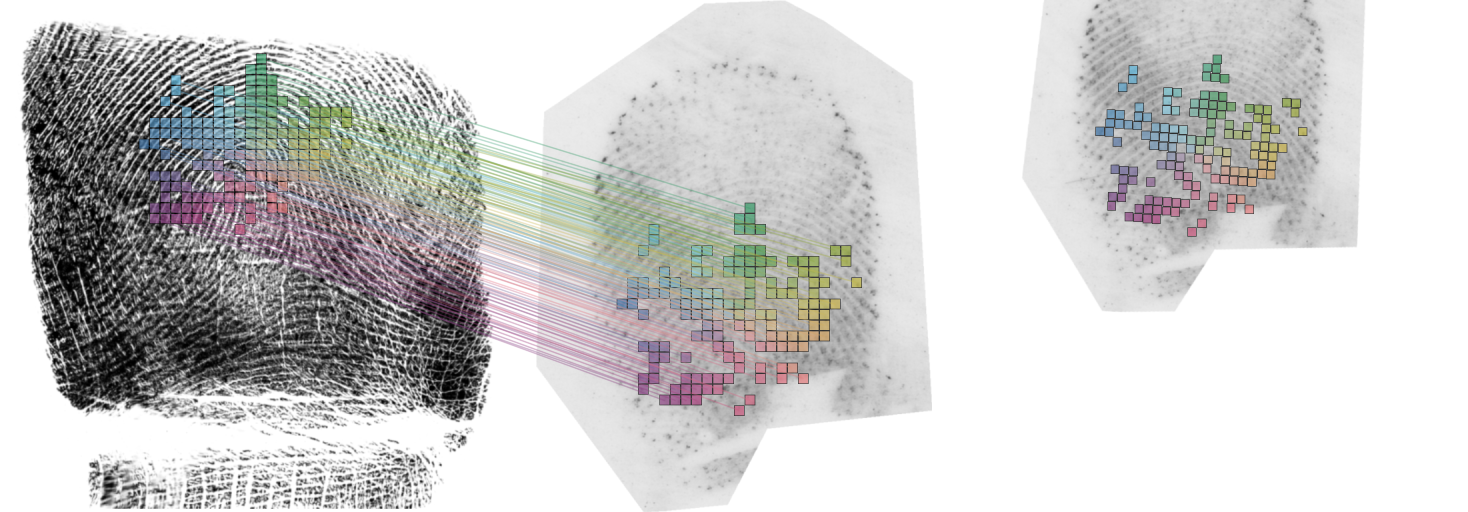}
    \caption{Cosine similarity: 0.670}
\end{subfigure}
\\[2pt]

\begin{subfigure}{\columnwidth}
    \includegraphics[width=\linewidth]{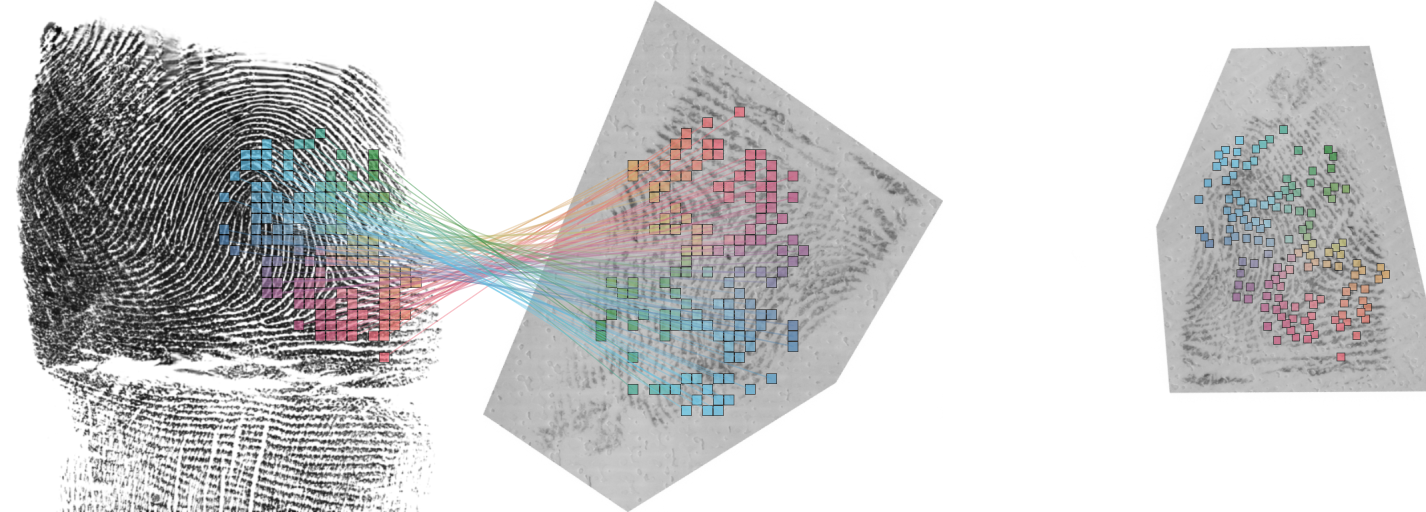}
    \caption{Cosine similarity: 0.495}
\end{subfigure}
\\[2pt]
\begin{subfigure}{\columnwidth}
    \includegraphics[width=\linewidth]{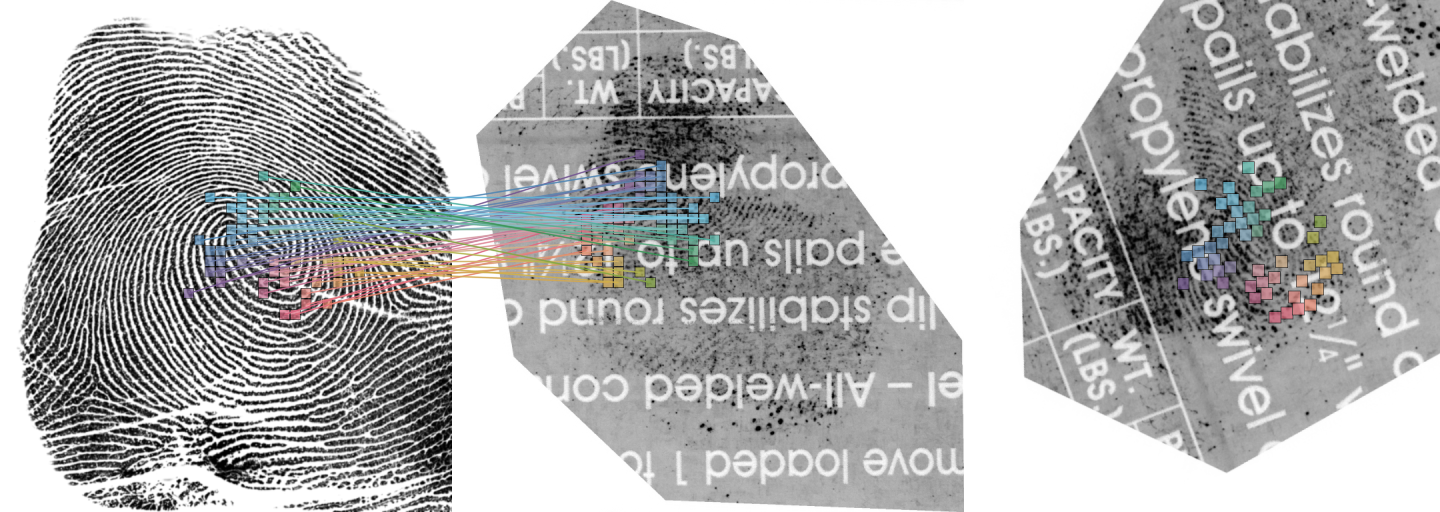}
    \caption{Cosine similarity: 0.421}
\end{subfigure}
\caption{Local descriptor correspondences from the AFID encoder. Exemplar (left), fingermark (centre), and the estimated similarity transform (right), which aligns the fingermark to the exemplar frame and is computed post-hoc from the correspondences purely for visualization.
}
\label{fig:local_rot_inv}
\end{figure}

The same invariance holds at the level of AFID's local descriptors. We extract local descriptors from the third stage of the encoder for a genuine fingerprint-fingermark pair, and match them by cosine similarity to obtain dense candidate correspondences. Running these through RANSAC retains the subset consistent with a single similarity transform, and applying that transform warps the fingermark into the exemplar's frame. Figure~\ref{fig:local_rot_inv} shows five such pairs: despite large rotation between print and mark, corresponding ridge regions retain high local similarity, and the recovered transform aligns the two impressions. The invariance learned at the level of the global embedding is thus also present in the local features from which it is composed. We stress that the transform is computed only for visualization and that the recognition pipeline uses no prior alignment. 

\section{Effect of the Background Regularizer on Embedding Norms}
\label{sup:bg_norms}

\begin{figure}[t]
  \centering
  \includegraphics[width=\columnwidth]{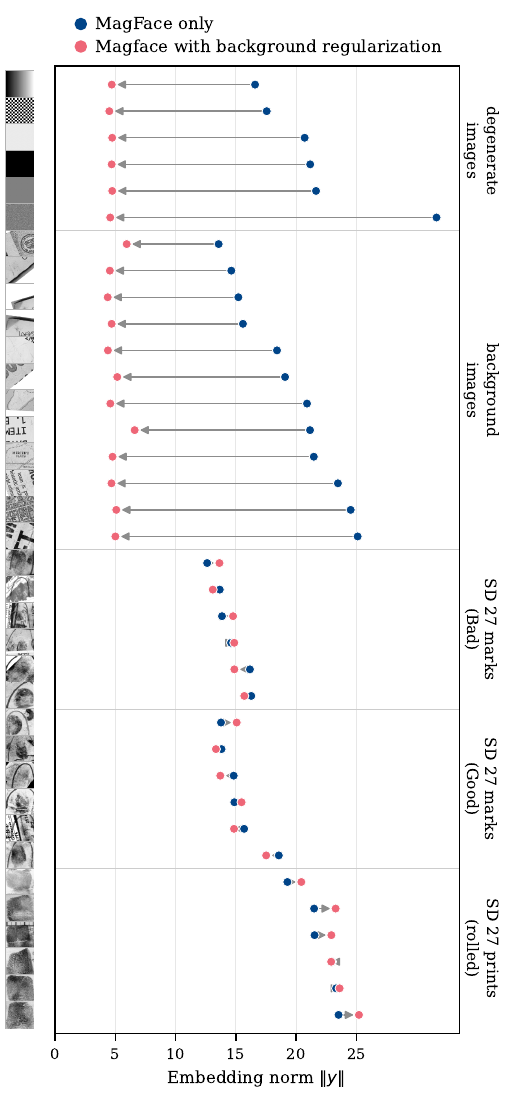}
  \caption{Effect of the background regularizer on embedding norms $\|\mathbf{y}\|$. We compare two models, one trained with MagFace loss only and the other trained by adding a background regularization term. While both models reach almost identical recognition performance, the latter produces much more stable norms on out of distribution samples. }
\label{fig:bg_norms}
\end{figure}

The background regularizer in Section~\ref{sec:recognition} is introduced to keep the embedding norm a reliable quality signal on inputs that carry no ridge content, which are never labelled and so are never seen by the MagFace objective. Figure~\ref{fig:bg_norms} shows its effect directly, comparing embedding norms $\|\mathbf{y}\|$ before and after adding the regularizer across five input categories: degenerate images (blank and synthetic fills), no-ridge background surfaces, and real SD~27 marks and prints of increasing quality.

Without the regularizer, degenerate and background inputs receive high norms, often comparable to those of genuine fingermarks. Since the norm serves as the quality signal $Q_n$, this is a failure mode: a blank surface or cluttered background would be scored the same as a high-quality impression. The regularizer corrects this, pulling the norms of degenerate and background inputs down toward the low-norm target $\tau = 5$ while leaving the norms of genuine marks and prints essentially unchanged. The ordering among real inputs is preserved, with rolled prints scoring highest and poor-quality marks lowest, so the regularizer removes the spurious high norms of no-ridge inputs without distorting the quality signal on real impressions. This confirms that it acts only where intended, and that it makes $Q_n$ usable as a quality measure across the full operational range rather than only on labeled samples. A useful side effect is that the clear norm gap between background and ridge-bearing inputs lets $\|\mathbf{y}\|$ double as a simple detector of whether an image contains visible friction ridges at all.

\section{Quality Prediction on Fingerprints}
\label{sup:fvc_quality}

The main-text quality evaluation in Section~\ref{sec:quality} focuses on fingermarks. For completeness we also assess it on the FVC fingerprint datasets. In the fingerprint domain, the established open quality assessment method is NFIQ~2~\cite{tabassi2021nfiq2}. Here we show EDC curves based on false non-match rate (FNMR) at fixed false match rate (FMR) of $10^{-3}$. This is the same operating point as in Section~\ref{sec:exp_verification}. As before, every experiment is repeated with three matchers (AFID, VeriFinger 2025.1, FLARE) to test cross-matcher generalization. We compare AFID-Q against VeriFinger's quality output and NFIQ~2, and summarize each curve by the normalized partial AUC (nAUC) over the first $20\%$ of discards.

\begin{figure*}[t]
  \centering
  \includegraphics[width=\textwidth]{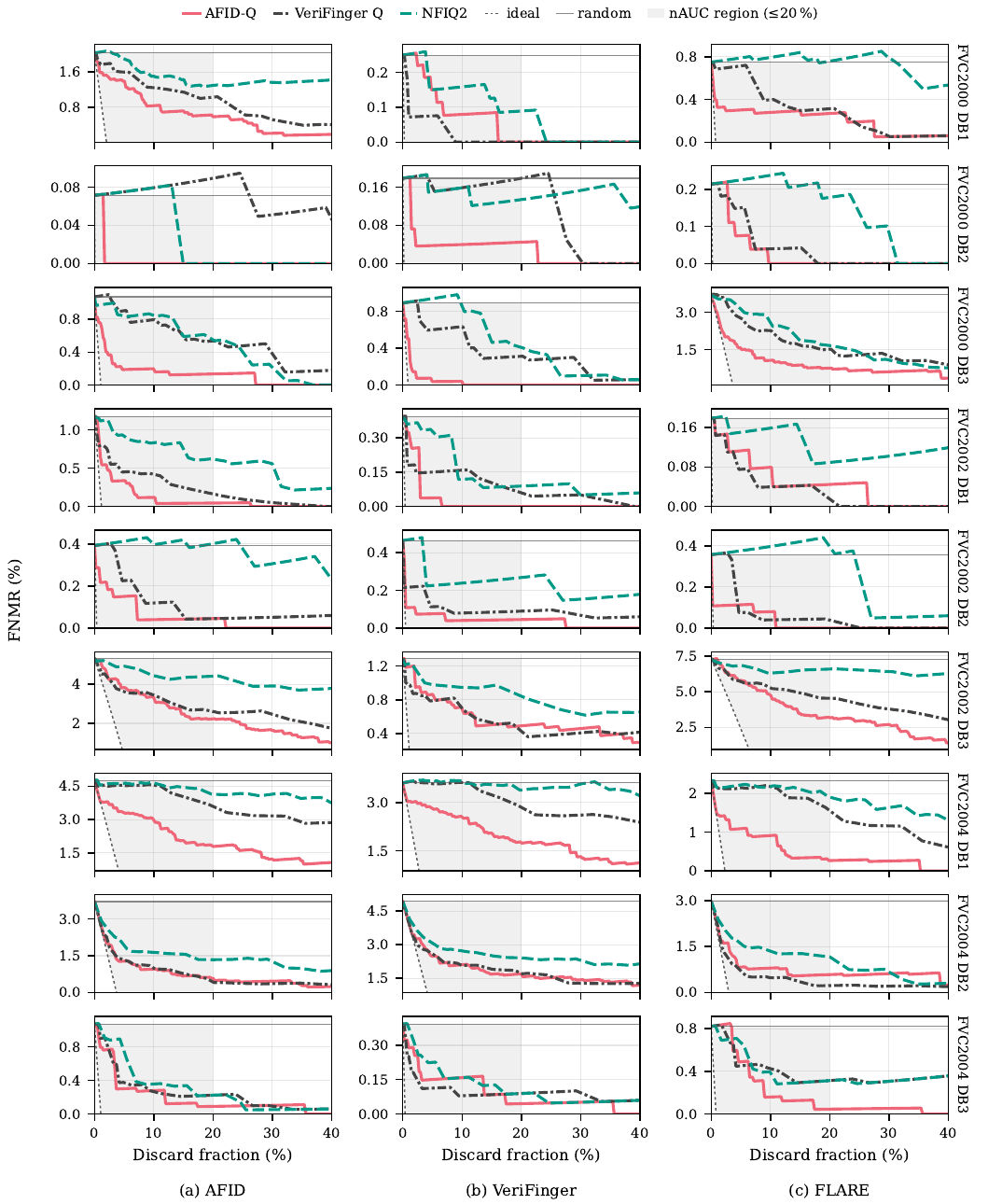}
  \caption{Verification EDC on the nine FVC databases (rows) for the three matchers (columns), at the FMR~$=10^{-3}$ operating point. Probes are discarded worst-quality-first; the curve reports FNMR over those remaining, and the shaded region marks the $\leq 20\%$ discard range over which nAUC is integrated. AFID-Q lies at or below the baselines in nearly every panel. Per-database values are omitted for legibility. Lower is better.}
  \label{fig:edc-fvc}
\end{figure*}

Figure~\ref{fig:edc-fvc} shows the EDC across all $27$ database\,$\times$\,matcher combinations. Averaged over the nine FVC databases, AFID-Q obtains the lowest nAUC under every matcher, with a mean of $0.321$ against $0.525$ for VeriFinger's quality value and $0.754$ for NFIQ~2. NFIQ~2 is the weakest of the three despite being purpose-built for fingerprints. That AFID-Q predicts fingerprint quality more accurately than the standardized fingerprint metric, while being trained primarily for fingermarks, indicates that the recognition-derived signal generalizes across the friction ridge quality range, instead of being predictive only to fingermark impressions.